\documentclass[letterpaper]{article} 
\usepackage{aaai2027}  
\usepackage[hyphens]{url}  
\usepackage{graphicx} 
\usepackage{natbib}  
\usepackage{caption} 
\usepackage{newfloat}
\usepackage{listings}
\DeclareCaptionStyle{ruled}{labelfont=normalfont,labelsep=colon,strut=off} 

\usepackage{booktabs}
\usepackage{multirow}
\usepackage{tabularx}
\usepackage{xcolor}
\usepackage{pgfplots}
\pgfplotsset{compat=1.18}
\usepackage{siunitx}
\usepackage{enumitem}
\usepackage{array}
\usepackage{amsmath,amssymb}
\newcolumntype{Y}{>{\raggedright\arraybackslash}X}

\definecolor{typeAblue}{RGB}{15,15,145}
\definecolor{typeAbackground}{RGB}{247,247,255}
\definecolor{typeBorange}{RGB}{166,82,0}
\definecolor{typeBbackground}{RGB}{255,249,242}

\newcommand{\examplequestionbox}[4]{%
\par\smallskip
\noindent
\begingroup
\setlength{\fboxsep}{0pt}%
\setlength{\fboxrule}{0.7pt}%
\fcolorbox{#1}{#2}{%
\parbox{\dimexpr\columnwidth-2\fboxrule\relax}{%
\setlength{\parindent}{0pt}%
\setlength{\parskip}{0pt}%
\setlength{\fboxsep}{3pt}%
\colorbox{#1}{%
\parbox{\dimexpr\linewidth-2\fboxsep\relax}{%
\color{white}\bfseries\footnotesize #3%
}}%
\par\smallskip
\begingroup
\footnotesize
\sloppy
\setlength{\leftskip}{4pt}%
\setlength{\rightskip}{4pt}%
#4
\par
\endgroup
\smallskip
}}%
\endgroup
\par\smallskip
}

\title{Findings of the First Teaching Monster Challenge: A Benchmark of Pedagogical Content Knowledge in AI Agents}
\author{
    Yi-Cheng Lin\textsuperscript{\rm 1},
    Yu-Kai Guo\equalcontrib\textsuperscript{\rm 1},
    Szu-Chi Chen\equalcontrib\textsuperscript{\rm 1},
    Bo-Han Feng\equalcontrib\textsuperscript{\rm 1},
    Yun-Man Hsu\equalcontrib\textsuperscript{\rm 1},
    Hsiang Hsieh\equalcontrib\textsuperscript{\rm 1},
    Yu-Jung Lin\equalcontrib\textsuperscript{\rm 1},
    Yue-Ling Wu\equalcontrib\textsuperscript{\rm 1},
    Jia-Kai Dong\equalcontrib\textsuperscript{\rm 1},
    An-Yu Cheng\textsuperscript{\rm 1},
    Yu-Han Huang\textsuperscript{\rm 1},
    Lok-Lam Ieong\textsuperscript{\rm 1},
    Kuan-Yu Chen\textsuperscript{\rm 1},
    Ming-Douo Tchouang\textsuperscript{\rm 1},
    Shao-Hua Sun\textsuperscript{\rm 1,\rm 2},
    Che Lin\textsuperscript{\rm 1},
    Jian-Jiun Ding\textsuperscript{\rm 1}, \\
    Hung-yi Lee\textsuperscript{\rm 1,\rm 2}
}
\affiliations{
    \textsuperscript{\rm 1}National Taiwan University\\
    \textsuperscript{\rm 2}NTU AI Center of Research Excellence
}

\begin{document}

\maketitle

\begin{abstract}
AI agents can now solve problems, answer like subject experts, and generate long-form multimodal content. 
However, whether they can adapt a lesson to fit a specified learner, which education calls Pedagogical Content Knowledge (PCK), has not been benchmarked.
To measure it, we introduce the Teaching Monster Challenge, the first instructional video generation benchmark to treat the learner persona as an explicit evaluation criterion. 
Each system is given a topic and a learner persona and must generate a complete instructional video.
Every video is screened by an LLM-judge, ranked by crowd pairwise voting, and finalized by an expert panel. 
The first edition shows that today's systems handle the content well but are far weaker at presenting it and adapting it to the learner. 
The same process exposes a limit of automatic judging. 
The LLM-judge separates a clear low-performing tail but ranks the strongest systems poorly. 
The strongest systems receive nearly identical scores from the judge, so its ranking of them does not match human preference. Progress therefore requires not only better teaching systems but also better automatic judges, and we release the benchmark, rubric, and human judgments as a testbed for both.
\end{abstract}

\section{Introduction}
\label{sec:introduction}
Around the world, most students are taught in groups from a single curriculum, even though instruction matched to each learner's level is far more effective~\citep{banerjee2016tarl}.
Tailoring instruction to every learner would require more teachers than currently exist, and the shortage is growing, with UNESCO estimating that 44 million more are needed by 2030~\citep{unesco2024teachers}.
AI agents have meanwhile grown proficient at subject-matter tasks, solving problems and answering questions like subject experts \citep{hendrycks2020measuring, cobbe2021training, chen2021codex} and generating long-form multimodal content.
If such an agent could teach, it could bring instruction fitted to each learner within reach at a scale human capacity cannot.

Teaching requires more than answering correctly. Shulman divides teacher knowledge into three kinds \citep{shulman1986}. 
Content knowledge is mastery of the subject matter itself, and pedagogical knowledge is the general craft of teaching that holds across subjects. 
Pedagogical Content Knowledge (PCK) is the blend of the two, the ability to transform a subject into a form a particular learner can grasp. 
Since knowing a subject does not guarantee being able to make a learner understand it, an agent's ability to teach depends not only on knowledge alone but also on whether it can adapt what it knows to a given learner.

This ability has not been benchmarked.
Existing benchmarks measure problem-solving accuracy \citep{hendrycks2020measuring, cobbe2021training, chen2021codex}, tutor responses in text dialogues \citep{tack2023bea, kochmar2025bea, macina2025mathtutorbench}, or perceptual video quality \citep{huang2024vbench, liu2024evalcrafter}, and none of them treats adaptation to a particular learner as an explicit criterion of evaluation.
A system can thus score well on all three and still remain untested on whether it can teach.
This gap is what the present paper sets out to fill.

We introduce the Teaching Monster Challenge\footnote{\url{https://teaching.monster/}}, the first instructional video generation benchmark to treat the learner persona as an explicit evaluation criterion.\footnote{The origin of the challenge's name is given in Appendix~\ref{app:appendix.orign.tm}.}
Given a course requirement and a learner persona, an agent must produce one complete instructional video from end to end with no human in the loop.
The task does not extend to interactive tutoring, where a teacher responds to a learner's questions and errors, which are too open-ended to compare across systems.
Because the input specifies which learner a video must fit, generic content is not enough, and this is what makes the challenge teaching rather than generation.

Submissions are scored on four evaluation dimensions.
\textit{Content Accuracy} measures factual correctness, \textit{Pedagogical Logic} measures coherent instructional sequencing, \textit{Learner Adaptability} measures adaptation to the learner, and \textit{Engagement and Multimodal Presentation} measures the quality of the video as a viewing experience.

The first edition is positioned to answer three questions about whether an agent can teach:
\begin{enumerate}[label=\textbf{RQ\arabic*.}, leftmargin=*, align=left, labelsep=0.5em]
    \item Which aspects of PCK do current systems already support well, and which remain challenging?
    \item Do systems adapt their instruction to the learner persona, and can people perceive that adaptation?
    \item Does the LLM-judge align with human judgment, and what explains their disagreements?
\end{enumerate}

This paper reports the first edition's findings, characterizing where the strongest systems meet expert pedagogical judgment and where they fall short.
The findings indicate that current systems are more mature at delivering content than at adapting it to the learner, and that the automated LLM-judge clearly separates weak submissions but performs poorly as a final ranker relative to human evaluation.
This paper also releases the first edition's assets, so that later work can compare systems on the same task and evaluation protocol.

\section{Related Work}

\paragraph{Document-to-Presentation Generation.}
Instructional video generation follows a pipeline of ingesting information, converting it into a visual modality, and presenting it to a learner.
Document-to-presentation generation shares this architectural philosophy, making it the most structurally analogous line of prior work.
DOC2PPT \citep{fu2022doc2ppt} introduced an early neural pipeline for this task; DocPres \citep{bandyopadhyay2024} and PPTAgent \citep{zheng2025pptagent} extend it with instruction-tuned LLMs.
Persona-Aware-D2S \citep{mondal2024persona} further conditions generation on pre-defined user classification.
These works take structured documents as input and slides or presentation videos as output, whereas our benchmark takes a course requirement and a learner persona as input and an instructional video as output.

\paragraph{Tutoring Benchmarks.}
Mainstream tutoring benchmarks evaluate pedagogical quality within text-based teacher-student dialogues.
BEA 2023 \citep{tack2023bea} frames tutoring evaluation as a generation task, scoring next-utterance generation using text-similarity metrics.
BEA 2025 \citep{kochmar2025bea} reframes the problem as classification, training separate classifiers for four pedagogical criteria on dialogue data and reporting F1 and accuracy.
MathTutorBench \citep{macina2025mathtutorbench} spans math expertise, student understanding, and open-ended scaffolding generation, combining accuracy-based metrics with a trained reward model to score response quality.
These benchmarks operate exclusively on text-based dialogues and do not extend to video.

\paragraph{Instructional Video Generation Benchmarks.}
Common video generation benchmarks mainly cover perceptual quality \citep{huang2024vbench,liu2024evalcrafter}.
Recent work has begun to incorporate instructional content into the evaluation.
TheoremExplainBench \citep{ku2025theoremexplain} evaluates generated videos across five dimensions, but does not address Learner Adaptability.
Paper2Video \citep{zhu2025paper2video} introduces PresentArena, an automated pairwise comparison protocol following the Arena paradigm \citep{chiang2024arena,jiang2024genaiarena}, using a VideoLLM as a proxy audience.
Code2Video \citep{chen2025code2video} introduces TeachQuiz, quantifying net knowledge transfer via VLM question answering and unlearning.
None of these works conditions on a specified learner.
To our knowledge, our benchmark is the first instructional video generation benchmark to treat learner persona as an explicit evaluation criterion.

\section{Task Design}
\label{sec:task-design}
\begin{figure*}[t]
    \centering
    \includegraphics[width=1.60\columnwidth]{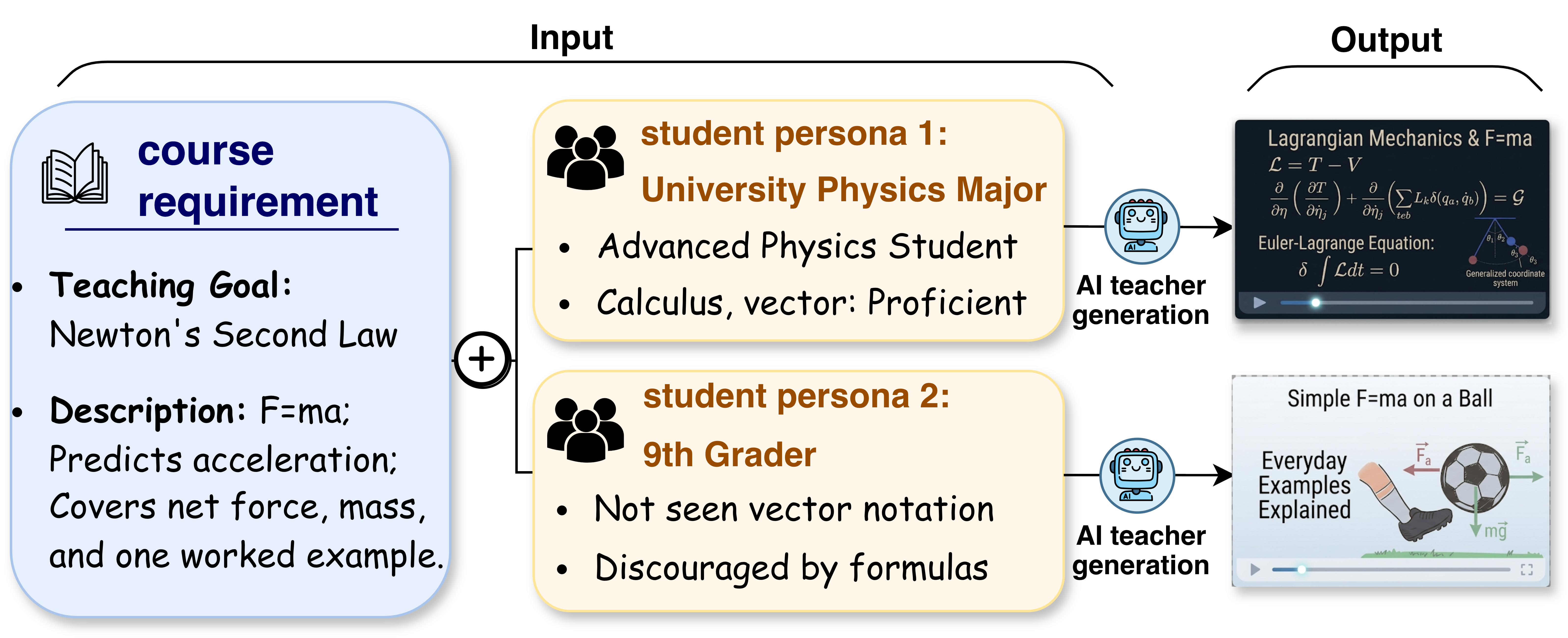}
    \caption{Task overview. A shared course requirement is paired with different student personas; systems must generate persona-adapted instructional videos, enabling comparison of whether a system adapts to the target learner.}
    \label{fig:overview}
\end{figure*}

Our challenge frames teaching as a single-shot task.
Given a learning need written in natural language, a system must return one complete instructional video, produced end-to-end with no human involved.
Asking for a complete video, rather than the isolated tutor turns of AI-tutoring benchmarks, forces the system to plan and build a whole lesson at once. 
That means breaking the concept down, ordering it, illustrating it, and delivering it across text, visuals, and narration.
Figure~\ref{fig:overview} summarizes the task flow.

\paragraph{Input.}
The benchmark is made up of items, and each item gives the system two inputs: a \texttt{course\_requirement} and a \texttt{learner\_persona}.
The \texttt{course\_requirement} states the learning objective and the key concepts the lesson should cover, and the \texttt{learner\_persona} describes, in free text, the learner's background, prior knowledge, and gaps.
Pairing the two is what makes the task teaching rather than generation, because the same material must be taught differently to different learners.
\paragraph{Scope and Matched-Pair Design.}
Items span four secondary-level STEM subjects (Physics, Biology, Computer Science, and Mathematics) and are anchored to the Advanced Placement curriculum \citep{collegeboard_ced}, a college-level secondary-school curriculum standard widely used in the United States.
A subset of items is organized into matched pairs that hold the \texttt{course\_requirement} fixed and vary the \texttt{learner\_persona}, so that a system's two videos for one pair differ only in the learner they target (Figure~\ref{fig:overview}).
Differencing within a pair removes item-level confounds such as topic difficulty and baseline production quality, so any remaining gap is attributable to the manipulated learner persona, yielding controlled evidence of adaptive teaching.

\section{Evaluation Framework}
\label{subsec:three-layer-protocol}
Measuring student learning gains directly from each video is infeasible at benchmark scale.  
We therefore evaluate teaching quality as a proxy for learning.
This follows established practice, where observer ratings of teaching quality predict achievement gains in classrooms~\cite{allen2013, kane2012met} and design-quality ratings of instructional videos relate to how much learners take from them~\cite{brame2016video}. 
Evaluation proceeds through a three-layer protocol: a Layer 1 LLM-judge screen over all submissions, a Layer 2 crowdsourced pairwise comparison over the shortlisted systems, and a Layer 3 expert ranking that determines the final challenge outcome.
Each layer narrows the field passed to the next, so that the more costly human evaluation of the later layers is concentrated on the strongest systems.
The following subsections describe each layer in turn.

\subsection{Layer 1: LLM-judge Screening}
\label{sec:llm-judge-sreening}

Layer 1 scores every submitted video with an automated LLM-judge and passes only the top systems on to human evaluation.
Scoring covers four dimensions, and the first three correspond to Shulman's three kinds of teacher knowledge~\citep{shulman1986}.
Each instructional video is scored on the following four evaluation dimensions.

\subsubsection{Content Accuracy}
Content Accuracy measures factual correctness and whether the claims in the exposition are free of hallucination.
As the most fundamental requirement of teaching, it captures the Subject Matter Content Knowledge that Shulman identifies among teacher knowledge, the foundation on which PCK is built.
\subsubsection{Pedagogical Logic}
Pedagogical Logic measures the scaffolding of the instruction, its coherence, and the clarity of the expository process.
This reflects pedagogical knowledge, another component of PCK.
\subsubsection{Learner Adaptability}
Learner Adaptability measures whether the instructional content matches the given learner persona, including the learner's grade level, attention span, and prior knowledge.
It operationalizes the adaptation of content to the specified learner, the transformation that is the core of PCK.
\subsubsection{Engagement and Multimodal Presentation}
Engagement and Multimodal Presentation measures the pacing of the narration, the alignment between the visuals and the speech, and whether instructional strategies such as analogies are used appropriately to elicit the learner's engagement.
It draws on Mayer's Cognitive Theory of Multimedia Learning.

\begin{figure}[ht]
    \centering
    \includegraphics[width=0.95\linewidth]{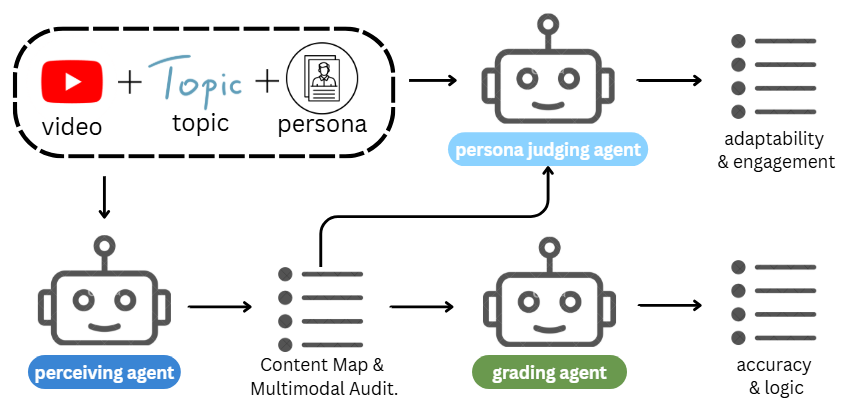}
    \caption{Data Flow for LLM-judge Scoring}
    \label{fig:rubric_dataflow}
\end{figure}

The scoring procedure conducted by the LLM-judge utilizes this three-agent architecture, dividing evaluation tasks among three collaborating agents: one for structured video perception, one for content and pedagogy scoring, and one for learner-persona scoring, as shown in Figure~\ref{fig:rubric_dataflow}.
The detailed scoring procedure and score computation are described in the Technical Appendix.
The four dimension scores select the top systems for Layer 2, so Layer 1 functions as a screening filter rather than the final arbiter of the challenge outcome.
Its fidelity is audited against the later human stages in Section~\ref{sec:rq3}.


\subsection{Layer 2: Crowdsourced Pairwise Comparison}

The highest-ranked systems from Layer 1 enter an Arena-style blind comparison\cite{chiang2024arena} study conducted with human raters recruited through a crowdsourcing platform.
For each comparison, the rater is shown the learner persona and course requirements, and is then asked to evaluate two instructional videos generated by two different systems for the same item to determine which one better helps the specified learner understand the topic.
This pairwise format avoids asking non-specialist raters to assign absolute scores on the full rubric, while still producing a preference signal suitable for system ranking.
The pairwise outcomes are aggregated into Elo ratings\cite{chiang2024arena} with bootstrap confidence intervals, and the resulting Layer 2 ranking selects the top systems for the final phase.

\subsection{Layer 3: Expert Ranking}

The top systems selected by Layer 2 generated instructional videos for expert-designed final-phase items.
These videos were evaluated by an expert panel composed of 10 secondary-school teachers, school principals, and university professors.
Each panel member reviewed a subset of the videos, watching them according to the descriptions of the same four evaluation dimensions and their sub-items, and ranked them directly without any instructions regarding score computation.
The aggregated expert ranking determines the final challenge outcome.
The expert layer also provides the reference signal used to interpret where the Layer 1 LLM-judge agrees with, or diverges from, human evaluation.


\section{Logistics}
\label{sec:logistics}

The Teaching Monster Challenge was organized into three competition phases: a Warm-up calibration phase, a Preliminary phase, and a Final phase. These phases defined the competition timeline, while ranked evaluation followed the three-layer protocol described in Section~\ref{subsec:three-layer-protocol}. The Warm-up phase allowed teams to validate their systems without affecting the ranking. 

The Preliminary phase served as the primary competitive pipeline. In this phase, Layer~1 was applied as an automated screening filter to evaluate the full field of submissions, forwarding only the highest-ranked systems to the Layer~2 crowdsourced evaluation. Together, these two layers selected the three finalists, and the Final phase subsequently used Layer~3 to determine the overall challenge outcome.

All phases operated through an organizer-built submission platform. 
Each participating team registered an API endpoint through which the platform delivered released items in the input format described in Section~\ref{sec:task-design}. 
Each API request enforced a 30-minute generation budget, and responses exceeding the budget were treated as failed submissions. 
Throughout the competition, the organizers maintained a Slack workspace for announcements, technical support, and the exchange of implementation experiences among participating teams.

The Warm-up calibration phase consisted of two rounds and did not contribute to the final ranking. The first round began on March 1, 2026, and used learning requests automatically generated by LLMs to stress-test API endpoints and validate the end-to-end submission pipeline. The second round began on April 14, 2026, and used committee-authored items. Across the two rounds, 46 teams submitted at least one instructional video, producing 1{,}696 videos in total. Together, the rounds allowed teams to test system reliability, familiarize themselves with the request format, and refine their pipelines using non-ranking feedback from the LLM-judge.

The Preliminary phase began on May 1, 2026, and consisted of two item-release rounds with 16 items each. Across the two rounds, 77 teams submitted at least one instructional video, producing 1{,}612 videos in total. 
In Layer 1, the LLM-judge, adapting the three-agent design of EduPanel~\citep{dong2026edupanel} and running on \texttt{gemini-3-flash}, assessed each submitted instructional video against a rubric of 22 items.
These items produced a score from 1 to 5 on each of the four evaluation dimensions, where a higher score is better.
The participating systems were ranked by the mean of their scores on four dimensions.
The 10 highest-ranked systems advanced to Layer 2.
In total, 59 raters recruited through Prolific contributed 246 comparisons within a fixed budget, and the outcomes were combined into Elo ratings with 90\% bootstrap confidence intervals.
The top three systems advanced to the Final phase; other ranking details are described in Appendix~\ref{app:layer2}.

The Final phase began on June 12, 2026, and determined the challenge outcome. The three finalist systems generated 48 instructional videos for 16 expert-designed items. These videos were evaluated in Layer 3 by an expert panel of secondary-school teachers, school principals, and university professors.
Each subject was covered by three panel members, two in the subject matter and one in pedagogy.
These experts watched every video in their subject and ranked the videos directly by the four evaluation dimensions, without giving numerical scores, using the evaluation interface shown in Appendix~\ref{app:layer3}.
Each system's average rank across the experts determined the final challenge standings. To enforce the no-human-in-the-loop requirement and support reproducibility, award-candidate teams were subsequently required to submit a complete system image, such as a Docker image, for post hoc verification of the end-to-end automated generation pipeline in an isolated environment.

\section{Systems}

\subsection{Submitted Systems}

A post-challenge questionnaire sent to all teams shows that nearly all submissions instantiate a six-stage cascade: persona parsing and routing, lesson planning, slide and script composition, visual rendering, narration synthesis, and assembly.
Recurring design choices include structuring the persona as JSON conditioning for every stage, difficulty-based model routing, web-search grounding of lesson plans, Bloom's-taxonomy constraints enforced by closed-loop LLM review, HTML/CSS slide templates with reviewer--fixer loops, split handling of conceptual (generated) versus mathematical (programmatic) figures, two-step planning for Manim animation, and forced alignment of narration to animation via word-level timestamps.
The Warm-up calibration and preliminary phases involved 46 and 77 teams, respectively.
We sent a questionnaire to all participating teams about their system architecture, and 15 teams responded.
We also reviewed every submitted video.
The following paragraphs are organized by the six stages, highlighting at each the notable design choices of the ten teams shortlisted in Layer 2.

\paragraph{Learner Persona Parsing and Routing.}
Eight teams out of the top 10 teams in Layer 2 first parse the free-text \texttt{learner\_persona} prompt into a structured persona profile in JSON format (grade level, prior knowledge, attention span) and use the structured persona as a condition in every following stage. 
Apart from that, one shortlisted team further analyzes what content is suitable for the corresponding educational level, deleting those out-of-scope topics that are too difficult for the learner persona. 

Aside from limiting the contents, some teams design a router that chooses the appropriate model for lesson planning depending on the educational level of the learner persona. 
For example, one team chooses the planning model according to the learner's attention span. 
For a persona with a normal attention span, it uses Claude Opus 4, which can produce more in-depth lessons. 
For a persona with a short attention span, it uses Claude Sonnet 4, whose shorter plans keep the lesson from running too long.

\paragraph{Lesson Planning \& Curriculum Blueprint.}
At this stage, the structured persona and the \texttt{course\_requirement} are passed to an LLM to generate a complete outline of the teaching slides. 
While most designs create a system prompt to instruct the LLM to collect information on the learning topic, the model is still prone to hallucination, producing out-of-scope or factually incorrect content. 
To reduce these errors, some teams ground the planning stage in retrieved sources, using web search through the Gemini API. 
This lowers the rate of factual errors, though it does not eliminate them.

On top of correctness, a good teaching video should also be well structured to effectively guide students to understand the concepts. 
As a result, some designs impose Bloom's taxonomy as a structural constraint, with a closed-loop LLM-judge check that verifies the outline against it.

\paragraph{Slide Composition \& Script Generation.}
After the outline is generated, slide composition and script generation can be derived from the outline page by page through another LLM agent. 
Slides are authored in various markup formats (Marp Markdown, LaTeX, HTML); among respondents in the Layer 2 shortlist, HTML/CSS was the most common choice, allowing richer layout control. 

In order to prevent the generated slides from having format issues or coding bugs, some teams constrain the generation to a fixed template library, reducing rendering errors; others add a review loop in which a checker agent reports layout and code issues back to the composition agent until the slide passes.

\paragraph{Visual \& Video Rendering.}
Images and animation are also key components of a good teaching slide. 
Although text-to-image models are widely available, respondents reported frequent hallucinations in images requiring factual precision (e.g., labeled diagrams). 
Therefore, some created an image library in advance for the agent to choose from. 
Others classify the needed images into conceptual and mathematical and only use AI-generated images for conceptual images while using matplotlib \cite{4160265} to generate mathematical images.

Manim \cite{sandersonManim} was the most common animation engine (4 of the top 10 respondents).
Because one-shot LLM generation of Manim code fails frequently, several teams split generation into a JSON storyboard step and a code-generation step, raising the rendering success rate.

\paragraph{Narration \& Audio Synthesis.}
Now that the slides and animation are ready, narration and other audio can be constructed by a text-to-speech model. 
At this stage, some teams map persona attributes to TTS parameters. For example, the speaking rate is slowed for younger learners, who need more time to process spoken explanations.
Although most of the teaching videos are presented in the voice of only one person, some teams adopt a two-voice design: one voice delivers the introduction and background, the other explains core concepts.

\paragraph{Video Composition \& Assembly.}
Lastly, the slides and the audio are combined to generate the final video. 
Some animations must fire exactly when the narration mentions them, which requires word-level timestamps. 
Top-ranked respondents recover word-level timestamps by running faster-whisper~\cite{systranFasterWhisper} (a reimplementation of Whisper~\cite{radford2023robust}) over the synthesized narration, effectively performing forced alignment, and triggering animations at the aligned words.

\subsection{Organizer baselines.}
Three organizer-built reference systems mark distinct regions of the design space. 
They are reference points rather than competition entries, and we score them with the same Layer 1 LLM-judge as the submitted systems so that the two can be read on a common scale. 
A \emph{cascaded pipeline} chains zero-shot commercial LLM APIs and open-weight local models through pedagogical planning, concurrent slide and narration generation, and a deterministic compilation phase in which word-level timestamps schedule an instructional pointer.
A \emph{commercial black-box generator} (Google NotebookLM Video Overview, accessed May 2026) synthesizes narrated instructional video end to end from the same two input fields.
A \emph{static retrieval framework} returns an existing Creative Commons video by ranking a frozen 257-video corpus with a cross-encoder~\cite{nogueira2019}, a model that reads the request and the candidate video together and scores how well the video's content matches the request, capturing genuine instructional relevance rather than surface keyword overlap. The learner persona is used only to keep the video within the learner's attention span, not to match its content. This baseline bounds what the reuse of existing content can achieve without any generation.
Module versions, configurations, and operational details for all three are given in Appendix~\ref{app:baselines}.



\begin{figure}[t]
\centering
\definecolor{auditplotteal}{RGB}{20,108,124}
\definecolor{auditplotci}{RGB}{48,52,56}
\definecolor{auditplotgrid}{RGB}{213,218,221}
\begin{tikzpicture}
\begin{axis}[
  width=0.48\textwidth,
  height=0.26\textwidth,
  xmin=0.5,
  xmax=77.5,
  ymin=0.8,
  ymax=5.2,
  xlabel={Team rank},
  ylabel={Mean LLM-judge score},
  xlabel style={font=\small},
  ylabel style={font=\small},
  tick label style={font=\footnotesize},
  xtick={1,8,15,22,29,36,43,50,57,64,71,77},
  ytick={1,2,3,4,5},
  ymajorgrids=true,
  grid style={draw=auditplotgrid, line width=0.55pt},
  axis lines*=left,
  axis line style={draw=black!80, line width=0.8pt},
  tick style={draw=black!80, line width=0.8pt},
  clip=true,
]
\draw[auditplotci, line width=0.65pt] (axis cs:1,4.585093) -- (axis cs:1,4.679595);
\draw[auditplotci, line width=0.65pt] (axis cs:0.88,4.585093) -- (axis cs:1.12,4.585093);
\draw[auditplotci, line width=0.65pt] (axis cs:0.88,4.679595) -- (axis cs:1.12,4.679595);
\draw[auditplotci, line width=0.65pt] (axis cs:2,4.586895) -- (axis cs:2,4.656386);
\draw[auditplotci, line width=0.65pt] (axis cs:1.88,4.586895) -- (axis cs:2.12,4.586895);
\draw[auditplotci, line width=0.65pt] (axis cs:1.88,4.656386) -- (axis cs:2.12,4.656386);
\draw[auditplotci, line width=0.65pt] (axis cs:3,4.535559) -- (axis cs:3,4.623035);
\draw[auditplotci, line width=0.65pt] (axis cs:2.88,4.535559) -- (axis cs:3.12,4.535559);
\draw[auditplotci, line width=0.65pt] (axis cs:2.88,4.623035) -- (axis cs:3.12,4.623035);
\draw[auditplotci, line width=0.65pt] (axis cs:4,4.524895) -- (axis cs:4,4.629324);
\draw[auditplotci, line width=0.65pt] (axis cs:3.88,4.524895) -- (axis cs:4.12,4.524895);
\draw[auditplotci, line width=0.65pt] (axis cs:3.88,4.629324) -- (axis cs:4.12,4.629324);
\draw[auditplotci, line width=0.65pt] (axis cs:5,4.509867) -- (axis cs:5,4.605289);
\draw[auditplotci, line width=0.65pt] (axis cs:4.88,4.509867) -- (axis cs:5.12,4.509867);
\draw[auditplotci, line width=0.65pt] (axis cs:4.88,4.605289) -- (axis cs:5.12,4.605289);
\draw[auditplotci, line width=0.65pt] (axis cs:6,4.500873) -- (axis cs:6,4.600690);
\draw[auditplotci, line width=0.65pt] (axis cs:5.88,4.500873) -- (axis cs:6.12,4.500873);
\draw[auditplotci, line width=0.65pt] (axis cs:5.88,4.600690) -- (axis cs:6.12,4.600690);
\draw[auditplotci, line width=0.65pt] (axis cs:7,4.490672) -- (axis cs:7,4.566515);
\draw[auditplotci, line width=0.65pt] (axis cs:6.88,4.490672) -- (axis cs:7.12,4.490672);
\draw[auditplotci, line width=0.65pt] (axis cs:6.88,4.566515) -- (axis cs:7.12,4.566515);
\draw[auditplotci, line width=0.65pt] (axis cs:8,4.421308) -- (axis cs:8,4.564161);
\draw[auditplotci, line width=0.65pt] (axis cs:7.88,4.421308) -- (axis cs:8.12,4.421308);
\draw[auditplotci, line width=0.65pt] (axis cs:7.88,4.564161) -- (axis cs:8.12,4.564161);
\draw[auditplotci, line width=0.65pt] (axis cs:9,4.384511) -- (axis cs:9,4.596426);
\draw[auditplotci, line width=0.65pt] (axis cs:8.88,4.384511) -- (axis cs:9.12,4.384511);
\draw[auditplotci, line width=0.65pt] (axis cs:8.88,4.596426) -- (axis cs:9.12,4.596426);
\draw[auditplotci, line width=0.65pt] (axis cs:10,4.379339) -- (axis cs:10,4.589567);
\draw[auditplotci, line width=0.65pt] (axis cs:9.88,4.379339) -- (axis cs:10.12,4.379339);
\draw[auditplotci, line width=0.65pt] (axis cs:9.88,4.589567) -- (axis cs:10.12,4.589567);
\draw[auditplotci, line width=0.65pt] (axis cs:11,4.428497) -- (axis cs:11,4.517753);
\draw[auditplotci, line width=0.65pt] (axis cs:10.88,4.428497) -- (axis cs:11.12,4.428497);
\draw[auditplotci, line width=0.65pt] (axis cs:10.88,4.517753) -- (axis cs:11.12,4.517753);
\draw[auditplotci, line width=0.65pt] (axis cs:12,4.397577) -- (axis cs:12,4.545692);
\draw[auditplotci, line width=0.65pt] (axis cs:11.88,4.397577) -- (axis cs:12.12,4.397577);
\draw[auditplotci, line width=0.65pt] (axis cs:11.88,4.545692) -- (axis cs:12.12,4.545692);
\draw[auditplotci, line width=0.65pt] (axis cs:13,4.362019) -- (axis cs:13,4.542512);
\draw[auditplotci, line width=0.65pt] (axis cs:12.88,4.362019) -- (axis cs:13.12,4.362019);
\draw[auditplotci, line width=0.65pt] (axis cs:12.88,4.542512) -- (axis cs:13.12,4.542512);
\draw[auditplotci, line width=0.65pt] (axis cs:14,4.374969) -- (axis cs:14,4.498313);
\draw[auditplotci, line width=0.65pt] (axis cs:13.88,4.374969) -- (axis cs:14.12,4.374969);
\draw[auditplotci, line width=0.65pt] (axis cs:13.88,4.498313) -- (axis cs:14.12,4.498313);
\draw[auditplotci, line width=0.65pt] (axis cs:15,4.375154) -- (axis cs:15,4.495314);
\draw[auditplotci, line width=0.65pt] (axis cs:14.88,4.375154) -- (axis cs:15.12,4.375154);
\draw[auditplotci, line width=0.65pt] (axis cs:14.88,4.495314) -- (axis cs:15.12,4.495314);
\draw[auditplotci, line width=0.65pt] (axis cs:16,4.233805) -- (axis cs:16,4.612445);
\draw[auditplotci, line width=0.65pt] (axis cs:15.88,4.233805) -- (axis cs:16.12,4.233805);
\draw[auditplotci, line width=0.65pt] (axis cs:15.88,4.612445) -- (axis cs:16.12,4.612445);
\draw[auditplotci, line width=0.65pt] (axis cs:17,4.323677) -- (axis cs:17,4.483198);
\draw[auditplotci, line width=0.65pt] (axis cs:16.88,4.323677) -- (axis cs:17.12,4.323677);
\draw[auditplotci, line width=0.65pt] (axis cs:16.88,4.483198) -- (axis cs:17.12,4.483198);
\draw[auditplotci, line width=0.65pt] (axis cs:18,4.319772) -- (axis cs:18,4.477103);
\draw[auditplotci, line width=0.65pt] (axis cs:17.88,4.319772) -- (axis cs:18.12,4.319772);
\draw[auditplotci, line width=0.65pt] (axis cs:17.88,4.477103) -- (axis cs:18.12,4.477103);
\draw[auditplotci, line width=0.65pt] (axis cs:19,3.982983) -- (axis cs:19,4.753128);
\draw[auditplotci, line width=0.65pt] (axis cs:18.88,3.982983) -- (axis cs:19.12,3.982983);
\draw[auditplotci, line width=0.65pt] (axis cs:18.88,4.753128) -- (axis cs:19.12,4.753128);
\draw[auditplotci, line width=0.65pt] (axis cs:20,4.156025) -- (axis cs:20,4.556319);
\draw[auditplotci, line width=0.65pt] (axis cs:19.88,4.156025) -- (axis cs:20.12,4.156025);
\draw[auditplotci, line width=0.65pt] (axis cs:19.88,4.556319) -- (axis cs:20.12,4.556319);
\draw[auditplotci, line width=0.65pt] (axis cs:21,4.284319) -- (axis cs:21,4.409587);
\draw[auditplotci, line width=0.65pt] (axis cs:20.88,4.284319) -- (axis cs:21.12,4.284319);
\draw[auditplotci, line width=0.65pt] (axis cs:20.88,4.409587) -- (axis cs:21.12,4.409587);
\draw[auditplotci, line width=0.65pt] (axis cs:22,4.212845) -- (axis cs:22,4.441892);
\draw[auditplotci, line width=0.65pt] (axis cs:21.88,4.212845) -- (axis cs:22.12,4.212845);
\draw[auditplotci, line width=0.65pt] (axis cs:21.88,4.441892) -- (axis cs:22.12,4.441892);
\draw[auditplotci, line width=0.65pt] (axis cs:23,4.106609) -- (axis cs:23,4.540744);
\draw[auditplotci, line width=0.65pt] (axis cs:22.88,4.106609) -- (axis cs:23.12,4.106609);
\draw[auditplotci, line width=0.65pt] (axis cs:22.88,4.540744) -- (axis cs:23.12,4.540744);
\draw[auditplotci, line width=0.65pt] (axis cs:24,4.198567) -- (axis cs:24,4.433308);
\draw[auditplotci, line width=0.65pt] (axis cs:23.88,4.198567) -- (axis cs:24.12,4.198567);
\draw[auditplotci, line width=0.65pt] (axis cs:23.88,4.433308) -- (axis cs:24.12,4.433308);
\draw[auditplotci, line width=0.65pt] (axis cs:25,4.231825) -- (axis cs:25,4.366144);
\draw[auditplotci, line width=0.65pt] (axis cs:24.88,4.231825) -- (axis cs:25.12,4.231825);
\draw[auditplotci, line width=0.65pt] (axis cs:24.88,4.366144) -- (axis cs:25.12,4.366144);
\draw[auditplotci, line width=0.65pt] (axis cs:26,4.015050) -- (axis cs:26,4.479424);
\draw[auditplotci, line width=0.65pt] (axis cs:25.88,4.015050) -- (axis cs:26.12,4.015050);
\draw[auditplotci, line width=0.65pt] (axis cs:25.88,4.479424) -- (axis cs:26.12,4.479424);
\draw[auditplotci, line width=0.65pt] (axis cs:27,4.033142) -- (axis cs:27,4.383396);
\draw[auditplotci, line width=0.65pt] (axis cs:26.88,4.033142) -- (axis cs:27.12,4.033142);
\draw[auditplotci, line width=0.65pt] (axis cs:26.88,4.383396) -- (axis cs:27.12,4.383396);
\draw[auditplotci, line width=0.65pt] (axis cs:28,3.926660) -- (axis cs:28,4.478340);
\draw[auditplotci, line width=0.65pt] (axis cs:27.88,3.926660) -- (axis cs:28.12,3.926660);
\draw[auditplotci, line width=0.65pt] (axis cs:27.88,4.478340) -- (axis cs:28.12,4.478340);
\draw[auditplotci, line width=0.65pt] (axis cs:29,4.005412) -- (axis cs:29,4.336150);
\draw[auditplotci, line width=0.65pt] (axis cs:28.88,4.005412) -- (axis cs:29.12,4.005412);
\draw[auditplotci, line width=0.65pt] (axis cs:28.88,4.336150) -- (axis cs:29.12,4.336150);
\draw[auditplotci, line width=0.65pt] (axis cs:30,3.909580) -- (axis cs:30,4.167451);
\draw[auditplotci, line width=0.65pt] (axis cs:29.88,3.909580) -- (axis cs:30.12,3.909580);
\draw[auditplotci, line width=0.65pt] (axis cs:29.88,4.167451) -- (axis cs:30.12,4.167451);
\draw[auditplotci, line width=0.65pt] (axis cs:31,3.887754) -- (axis cs:31,4.107558);
\draw[auditplotci, line width=0.65pt] (axis cs:30.88,3.887754) -- (axis cs:31.12,3.887754);
\draw[auditplotci, line width=0.65pt] (axis cs:30.88,4.107558) -- (axis cs:31.12,4.107558);
\draw[auditplotci, line width=0.65pt] (axis cs:32,3.711035) -- (axis cs:32,4.203965);
\draw[auditplotci, line width=0.65pt] (axis cs:31.88,3.711035) -- (axis cs:32.12,3.711035);
\draw[auditplotci, line width=0.65pt] (axis cs:31.88,4.203965) -- (axis cs:32.12,4.203965);
\draw[auditplotci, line width=0.65pt] (axis cs:33,3.728296) -- (axis cs:33,3.908891);
\draw[auditplotci, line width=0.65pt] (axis cs:32.88,3.728296) -- (axis cs:33.12,3.728296);
\draw[auditplotci, line width=0.65pt] (axis cs:32.88,3.908891) -- (axis cs:33.12,3.908891);
\draw[auditplotci, line width=0.65pt] (axis cs:34,3.627928) -- (axis cs:34,3.973322);
\draw[auditplotci, line width=0.65pt] (axis cs:33.88,3.627928) -- (axis cs:34.12,3.627928);
\draw[auditplotci, line width=0.65pt] (axis cs:33.88,3.973322) -- (axis cs:34.12,3.973322);
\draw[auditplotci, line width=0.65pt] (axis cs:35,2.787061) -- (axis cs:35,4.553939);
\draw[auditplotci, line width=0.65pt] (axis cs:34.88,2.787061) -- (axis cs:35.12,2.787061);
\draw[auditplotci, line width=0.65pt] (axis cs:34.88,4.553939) -- (axis cs:35.12,4.553939);
\draw[auditplotci, line width=0.65pt] (axis cs:36,3.236349) -- (axis cs:36,4.015839);
\draw[auditplotci, line width=0.65pt] (axis cs:35.88,3.236349) -- (axis cs:36.12,3.236349);
\draw[auditplotci, line width=0.65pt] (axis cs:35.88,4.015839) -- (axis cs:36.12,4.015839);
\draw[auditplotci, line width=0.65pt] (axis cs:38,1.534807) -- (axis cs:38,5.680193);
\draw[auditplotci, line width=0.65pt] (axis cs:37.88,1.534807) -- (axis cs:38.12,1.534807);
\draw[auditplotci, line width=0.65pt] (axis cs:37.88,5.680193) -- (axis cs:38.12,5.680193);
\draw[auditplotci, line width=0.65pt] (axis cs:39,3.187891) -- (axis cs:39,4.023359);
\draw[auditplotci, line width=0.65pt] (axis cs:38.88,3.187891) -- (axis cs:39.12,3.187891);
\draw[auditplotci, line width=0.65pt] (axis cs:38.88,4.023359) -- (axis cs:39.12,4.023359);
\draw[auditplotci, line width=0.65pt] (axis cs:40,3.370076) -- (axis cs:40,3.797424);
\draw[auditplotci, line width=0.65pt] (axis cs:39.88,3.370076) -- (axis cs:40.12,3.370076);
\draw[auditplotci, line width=0.65pt] (axis cs:39.88,3.797424) -- (axis cs:40.12,3.797424);
\draw[auditplotci, line width=0.65pt] (axis cs:41,2.408070) -- (axis cs:41,4.723596);
\draw[auditplotci, line width=0.65pt] (axis cs:40.88,2.408070) -- (axis cs:41.12,2.408070);
\draw[auditplotci, line width=0.65pt] (axis cs:40.88,4.723596) -- (axis cs:41.12,4.723596);
\draw[auditplotci, line width=0.65pt] (axis cs:42,3.220733) -- (axis cs:42,3.808954);
\draw[auditplotci, line width=0.65pt] (axis cs:41.88,3.220733) -- (axis cs:42.12,3.220733);
\draw[auditplotci, line width=0.65pt] (axis cs:41.88,3.808954) -- (axis cs:42.12,3.808954);
\draw[auditplotci, line width=0.65pt] (axis cs:43,2.679388) -- (axis cs:43,4.214362);
\draw[auditplotci, line width=0.65pt] (axis cs:42.88,2.679388) -- (axis cs:43.12,2.679388);
\draw[auditplotci, line width=0.65pt] (axis cs:42.88,4.214362) -- (axis cs:43.12,4.214362);
\draw[auditplotci, line width=0.65pt] (axis cs:44,3.204384) -- (axis cs:44,3.685616);
\draw[auditplotci, line width=0.65pt] (axis cs:43.88,3.204384) -- (axis cs:44.12,3.204384);
\draw[auditplotci, line width=0.65pt] (axis cs:43.88,3.685616) -- (axis cs:44.12,3.685616);
\draw[auditplotci, line width=0.65pt] (axis cs:45,2.999594) -- (axis cs:45,3.871031);
\draw[auditplotci, line width=0.65pt] (axis cs:44.88,2.999594) -- (axis cs:45.12,2.999594);
\draw[auditplotci, line width=0.65pt] (axis cs:44.88,3.871031) -- (axis cs:45.12,3.871031);
\draw[auditplotci, line width=0.65pt] (axis cs:46,-2.644120) -- (axis cs:46,9.109120);
\draw[auditplotci, line width=0.65pt] (axis cs:45.88,-2.644120) -- (axis cs:46.12,-2.644120);
\draw[auditplotci, line width=0.65pt] (axis cs:45.88,9.109120) -- (axis cs:46.12,9.109120);
\draw[auditplotci, line width=0.65pt] (axis cs:47,2.907198) -- (axis cs:47,3.346969);
\draw[auditplotci, line width=0.65pt] (axis cs:46.88,2.907198) -- (axis cs:47.12,2.907198);
\draw[auditplotci, line width=0.65pt] (axis cs:46.88,3.346969) -- (axis cs:47.12,3.346969);
\draw[auditplotci, line width=0.65pt] (axis cs:48,2.604775) -- (axis cs:48,3.386631);
\draw[auditplotci, line width=0.65pt] (axis cs:47.88,2.604775) -- (axis cs:48.12,2.604775);
\draw[auditplotci, line width=0.65pt] (axis cs:47.88,3.386631) -- (axis cs:48.12,3.386631);
\draw[auditplotci, line width=0.65pt] (axis cs:49,2.625028) -- (axis cs:49,3.302941);
\draw[auditplotci, line width=0.65pt] (axis cs:48.88,2.625028) -- (axis cs:49.12,2.625028);
\draw[auditplotci, line width=0.65pt] (axis cs:48.88,3.302941) -- (axis cs:49.12,3.302941);
\draw[auditplotci, line width=0.65pt] (axis cs:50,2.713451) -- (axis cs:50,3.189883);
\draw[auditplotci, line width=0.65pt] (axis cs:49.88,2.713451) -- (axis cs:50.12,2.713451);
\draw[auditplotci, line width=0.65pt] (axis cs:49.88,3.189883) -- (axis cs:50.12,3.189883);
\draw[auditplotci, line width=0.65pt] (axis cs:51,2.397737) -- (axis cs:51,3.444322);
\draw[auditplotci, line width=0.65pt] (axis cs:50.88,2.397737) -- (axis cs:51.12,2.397737);
\draw[auditplotci, line width=0.65pt] (axis cs:50.88,3.444322) -- (axis cs:51.12,3.444322);
\draw[auditplotci, line width=0.65pt] (axis cs:52,2.497286) -- (axis cs:52,3.310527);
\draw[auditplotci, line width=0.65pt] (axis cs:51.88,2.497286) -- (axis cs:52.12,2.497286);
\draw[auditplotci, line width=0.65pt] (axis cs:51.88,3.310527) -- (axis cs:52.12,3.310527);
\draw[auditplotci, line width=0.65pt] (axis cs:53,2.385106) -- (axis cs:53,3.249581);
\draw[auditplotci, line width=0.65pt] (axis cs:52.88,2.385106) -- (axis cs:53.12,2.385106);
\draw[auditplotci, line width=0.65pt] (axis cs:52.88,3.249581) -- (axis cs:53.12,3.249581);
\draw[auditplotci, line width=0.65pt] (axis cs:54,2.275055) -- (axis cs:54,3.332445);
\draw[auditplotci, line width=0.65pt] (axis cs:53.88,2.275055) -- (axis cs:54.12,2.275055);
\draw[auditplotci, line width=0.65pt] (axis cs:53.88,3.332445) -- (axis cs:54.12,3.332445);
\draw[auditplotci, line width=0.65pt] (axis cs:55,1.550279) -- (axis cs:55,3.999721);
\draw[auditplotci, line width=0.65pt] (axis cs:54.88,1.550279) -- (axis cs:55.12,1.550279);
\draw[auditplotci, line width=0.65pt] (axis cs:54.88,3.999721) -- (axis cs:55.12,3.999721);
\draw[auditplotci, line width=0.65pt] (axis cs:56,2.396632) -- (axis cs:56,3.036917);
\draw[auditplotci, line width=0.65pt] (axis cs:55.88,2.396632) -- (axis cs:56.12,2.396632);
\draw[auditplotci, line width=0.65pt] (axis cs:55.88,3.036917) -- (axis cs:56.12,3.036917);
\draw[auditplotci, line width=0.65pt] (axis cs:57,1.304309) -- (axis cs:57,4.039024);
\draw[auditplotci, line width=0.65pt] (axis cs:56.88,1.304309) -- (axis cs:57.12,1.304309);
\draw[auditplotci, line width=0.65pt] (axis cs:56.88,4.039024) -- (axis cs:57.12,4.039024);
\draw[auditplotci, line width=0.65pt] (axis cs:58,2.011760) -- (axis cs:58,3.106990);
\draw[auditplotci, line width=0.65pt] (axis cs:57.88,2.011760) -- (axis cs:58.12,2.011760);
\draw[auditplotci, line width=0.65pt] (axis cs:57.88,3.106990) -- (axis cs:58.12,3.106990);
\draw[auditplotci, line width=0.65pt] (axis cs:59,2.480184) -- (axis cs:59,2.619816);
\draw[auditplotci, line width=0.65pt] (axis cs:58.88,2.480184) -- (axis cs:59.12,2.480184);
\draw[auditplotci, line width=0.65pt] (axis cs:58.88,2.619816) -- (axis cs:59.12,2.619816);
\draw[auditplotci, line width=0.65pt] (axis cs:60,1.946014) -- (axis cs:60,3.144541);
\draw[auditplotci, line width=0.65pt] (axis cs:59.88,1.946014) -- (axis cs:60.12,1.946014);
\draw[auditplotci, line width=0.65pt] (axis cs:59.88,3.144541) -- (axis cs:60.12,3.144541);
\draw[auditplotci, line width=0.65pt] (axis cs:61,2.283583) -- (axis cs:61,2.724195);
\draw[auditplotci, line width=0.65pt] (axis cs:60.88,2.283583) -- (axis cs:61.12,2.283583);
\draw[auditplotci, line width=0.65pt] (axis cs:60.88,2.724195) -- (axis cs:61.12,2.724195);
\draw[auditplotci, line width=0.65pt] (axis cs:62,1.988812) -- (axis cs:62,2.857751);
\draw[auditplotci, line width=0.65pt] (axis cs:61.88,1.988812) -- (axis cs:62.12,1.988812);
\draw[auditplotci, line width=0.65pt] (axis cs:61.88,2.857751) -- (axis cs:62.12,2.857751);
\draw[auditplotci, line width=0.65pt] (axis cs:63,1.883196) -- (axis cs:63,2.923992);
\draw[auditplotci, line width=0.65pt] (axis cs:62.88,1.883196) -- (axis cs:63.12,1.883196);
\draw[auditplotci, line width=0.65pt] (axis cs:62.88,2.923992) -- (axis cs:63.12,2.923992);
\draw[auditplotci, line width=0.65pt] (axis cs:64,1.871541) -- (axis cs:64,2.814396);
\draw[auditplotci, line width=0.65pt] (axis cs:63.88,1.871541) -- (axis cs:64.12,1.871541);
\draw[auditplotci, line width=0.65pt] (axis cs:63.88,2.814396) -- (axis cs:64.12,2.814396);
\draw[auditplotci, line width=0.65pt] (axis cs:65,1.606589) -- (axis cs:65,3.061744);
\draw[auditplotci, line width=0.65pt] (axis cs:64.88,1.606589) -- (axis cs:65.12,1.606589);
\draw[auditplotci, line width=0.65pt] (axis cs:64.88,3.061744) -- (axis cs:65.12,3.061744);
\draw[auditplotci, line width=0.65pt] (axis cs:66,1.938635) -- (axis cs:66,2.696209);
\draw[auditplotci, line width=0.65pt] (axis cs:65.88,1.938635) -- (axis cs:66.12,1.938635);
\draw[auditplotci, line width=0.65pt] (axis cs:65.88,2.696209) -- (axis cs:66.12,2.696209);
\draw[auditplotci, line width=0.65pt] (axis cs:67,1.651197) -- (axis cs:67,2.766928);
\draw[auditplotci, line width=0.65pt] (axis cs:66.88,1.651197) -- (axis cs:67.12,1.651197);
\draw[auditplotci, line width=0.65pt] (axis cs:66.88,2.766928) -- (axis cs:67.12,2.766928);
\draw[auditplotci, line width=0.65pt] (axis cs:68,1.614288) -- (axis cs:68,2.528056);
\draw[auditplotci, line width=0.65pt] (axis cs:67.88,1.614288) -- (axis cs:68.12,1.614288);
\draw[auditplotci, line width=0.65pt] (axis cs:67.88,2.528056) -- (axis cs:68.12,2.528056);
\draw[auditplotci, line width=0.65pt] (axis cs:69,1.729622) -- (axis cs:69,2.388815);
\draw[auditplotci, line width=0.65pt] (axis cs:68.88,1.729622) -- (axis cs:69.12,1.729622);
\draw[auditplotci, line width=0.65pt] (axis cs:68.88,2.388815) -- (axis cs:69.12,2.388815);
\draw[auditplotci, line width=0.65pt] (axis cs:70,1.646023) -- (axis cs:70,2.272102);
\draw[auditplotci, line width=0.65pt] (axis cs:69.88,1.646023) -- (axis cs:70.12,1.646023);
\draw[auditplotci, line width=0.65pt] (axis cs:69.88,2.272102) -- (axis cs:70.12,2.272102);
\draw[auditplotci, line width=0.65pt] (axis cs:71,1.671714) -- (axis cs:71,1.976411);
\draw[auditplotci, line width=0.65pt] (axis cs:70.88,1.671714) -- (axis cs:71.12,1.671714);
\draw[auditplotci, line width=0.65pt] (axis cs:70.88,1.976411) -- (axis cs:71.12,1.976411);
\draw[auditplotci, line width=0.65pt] (axis cs:73,0.981057) -- (axis cs:73,1.179100);
\draw[auditplotci, line width=0.65pt] (axis cs:72.88,0.981057) -- (axis cs:73.12,0.981057);
\draw[auditplotci, line width=0.65pt] (axis cs:72.88,1.179100) -- (axis cs:73.12,1.179100);
\draw[auditplotci, line width=0.65pt] (axis cs:74,0.995486) -- (axis cs:74,1.046701);
\draw[auditplotci, line width=0.65pt] (axis cs:73.88,0.995486) -- (axis cs:74.12,0.995486);
\draw[auditplotci, line width=0.65pt] (axis cs:73.88,1.046701) -- (axis cs:74.12,1.046701);
\draw[auditplotci, line width=0.65pt] (axis cs:75,0.996123) -- (axis cs:75,1.021846);
\draw[auditplotci, line width=0.65pt] (axis cs:74.88,0.996123) -- (axis cs:75.12,0.996123);
\draw[auditplotci, line width=0.65pt] (axis cs:74.88,1.021846) -- (axis cs:75.12,1.021846);
\draw[auditplotci, line width=0.65pt] (axis cs:76,1.000000) -- (axis cs:76,1.000000);
\draw[auditplotci, line width=0.65pt] (axis cs:75.88,1.000000) -- (axis cs:76.12,1.000000);
\draw[auditplotci, line width=0.65pt] (axis cs:75.88,1.000000) -- (axis cs:76.12,1.000000);
\draw[auditplotci, line width=0.65pt] (axis cs:77,1.000000) -- (axis cs:77,1.000000);
\draw[auditplotci, line width=0.65pt] (axis cs:76.88,1.000000) -- (axis cs:77.12,1.000000);
\draw[auditplotci, line width=0.65pt] (axis cs:76.88,1.000000) -- (axis cs:77.12,1.000000);
\addplot+[
  color=auditplotteal,
  line width=0.9pt,
  no marks
] coordinates {
  (1,4.632344) (2,4.621641) (3,4.579297) (4,4.577109) (5,4.557578) (6,4.550781) (7,4.528594) (8,4.492734) (9,4.490469) (10,4.484453) (11,4.473125) (12,4.471635) (13,4.452266) (14,4.436641) (15,4.435234) (16,4.423125) (17,4.403437) (18,4.398438) (19,4.368056) (20,4.356172) (21,4.346953) (22,4.327368) (23,4.323676) (24,4.315938) (25,4.298984) (26,4.247237) (27,4.208269) (28,4.202500) (29,4.170781) (30,4.038516) (31,3.997656) (32,3.957500) (33,3.818594) (34,3.800625) (35,3.670500) (36,3.626094) (37,3.612500) (38,3.607500) (39,3.605625) (40,3.583750) (41,3.565833) (42,3.514844) (43,3.446875) (44,3.445000) (45,3.435313) (46,3.232500) (47,3.127083) (48,2.995703) (49,2.963984) (50,2.951667) (51,2.921029) (52,2.903906) (53,2.817344) (54,2.803750) (55,2.775000) (56,2.716774) (57,2.671667) (58,2.559375) (59,2.550000) (60,2.545278) (61,2.503889) (62,2.423281) (63,2.403594) (64,2.342969) (65,2.334167) (66,2.317422) (67,2.209062) (68,2.071172) (69,2.059219) (70,1.959062) (71,1.824062) (72,1.175000) (73,1.080078) (74,1.021094) (75,1.008984) (76,1.000000) (77,1.000000)
};
\addplot+[
  only marks,
  color=auditplotteal,
  mark=*,
  mark size=0.8pt,
  mark options={
    draw=auditplotteal,
    fill=white,
    line width=0.5pt
  }
] coordinates {
  (1,4.632344) (2,4.621641) (3,4.579297) (4,4.577109) (5,4.557578) (6,4.550781) (7,4.528594) (8,4.492734) (9,4.490469) (10,4.484453) (11,4.473125) (12,4.471635) (13,4.452266) (14,4.436641) (15,4.435234) (16,4.423125) (17,4.403437) (18,4.398438) (19,4.368056) (20,4.356172) (21,4.346953) (22,4.327368) (23,4.323676) (24,4.315938) (25,4.298984) (26,4.247237) (27,4.208269) (28,4.202500) (29,4.170781) (30,4.038516) (31,3.997656) (32,3.957500) (33,3.818594) (34,3.800625) (35,3.670500) (36,3.626094) (37,3.612500) (38,3.607500) (39,3.605625) (40,3.583750) (41,3.565833) (42,3.514844) (43,3.446875) (44,3.445000) (45,3.435313) (46,3.232500) (47,3.127083) (48,2.995703) (49,2.963984) (50,2.951667) (51,2.921029) (52,2.903906) (53,2.817344) (54,2.803750) (55,2.775000) (56,2.716774) (57,2.671667) (58,2.559375) (59,2.550000) (60,2.545278) (61,2.503889) (62,2.423281) (63,2.403594) (64,2.342969) (65,2.334167) (66,2.317422) (67,2.209062) (68,2.071172) (69,2.059219) (70,1.959062) (71,1.824062) (72,1.175000) (73,1.080078) (74,1.021094) (75,1.008984) (76,1.000000) (77,1.000000)
};
\end{axis}
\end{tikzpicture}
\caption{LLM-judge scores for all submitted systems. Each point shows one system's mean score over its preliminary-round videos, ordered by rank, with 95\% confidence intervals.}
\label{fig:score-decline-by-rank}
\end{figure}
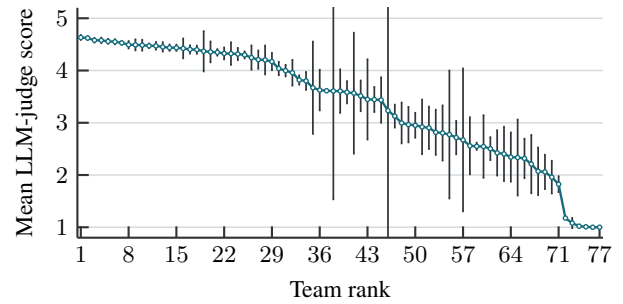

\section{Results}
\label{sec:results}
 We first report the aggregate challenge results in Sec.~\ref{sec:overall} and then examine them in greater detail. We analyze the pedagogical strengths and weaknesses revealed by the Layer 1 rubric (RQ1), investigate learner-persona adaptation (RQ2), and finally compare the Layer 1 judge with human evaluation (RQ3).
\subsection{Overall Challenge Results}
\label{sec:overall}

Table~\ref{tab:main} reports all three evaluation layers, from 77 submitting teams through the ten-system shortlist to the three finalists.
Figure~\ref{fig:score-decline-by-rank} shows a long tail of weak or off-topic submissions whose scores fall well below the others.
Removing this tail let the costly human evaluation in Layers 2 and 3 focus on the ten shortlisted systems instead of every team.
Among these ten, the crowd ranked \texttt{tsunumon} first but could not order \texttt{OmegaZeroRe} and \texttt{ScaffoldAI}, whose Elo intervals overlapped.
The Layer~3 experts settled that order, and their rankings agreed at Kendall's $W = 0.41$.
\texttt{tsunumon} won the challenge, followed by \texttt{OmegaZeroRe} and \texttt{ScaffoldAI}.

Table~\ref{tab:main} also scores the three organizer baselines, which the judge and the human raters see very differently.
The judge gives the static retrieval framework the lowest scores of all, mainly on Content Accuracy and Pedagogical Logic.
Its deduction records show that most penalties come from videos missing or only partly covering the requested topic, not from factual errors.
The low scores measure how well retrieval matched the item, not how well the videos teach.
Table~\ref{tab:baseline-human} then compares each baseline against the first-, second-, and fifth-ranked systems under the Layer 2 protocol.
All three placed between first and second.
The judge's content scores therefore do not track what raters reward, a gap Section~\ref{sec:rq3} examines.
Only the challenge winner \texttt{tsunumon} beat the baselines, so the systems below it have not yet clearly improved on a mature commercial product or a well-chosen human-made video.
 
\begin{table*}[t]
\centering
\small
\caption{Results of all three evaluation layers, with the ten shortlisted systems ordered by Layer~2 rank and the organizer baselines below. The dimension columns abbreviate Content Accuracy, Pedagogical Logic, Learner Adaptability, Engagement, and Multimodal Presentation. Elo entries carry 90\% bootstrap confidence intervals, and ``--'' means not ranked at that stage.}
\label{tab:main}
\setlength{\tabcolsep}{5pt}
\begin{tabular}{lccccc c cc c}
\toprule
 & \multicolumn{5}{c}{Layer 1 (LLM judge)} & & \multicolumn{2}{c}{Layer 2 (crowd)} & Layer 3 \\
\cmidrule{2-6}\cmidrule{8-9}\cmidrule{10-10}
System & Acc. & Logic & Adapt. & Eng. & Rank & & Elo (90\% CI) & Rank & Rank \\
\midrule
tsunumon             & 4.72 & 4.66 & 4.48 & 4.46 & 4  & & 1670 (1588--1742) & 1  & 1 \\
ScaffoldAI           & 4.72 & 4.66 & 4.47 & 4.64 & 3  & & 1570 (1459--1669) & 2  & 3 \\
OmegaZeroRe          & 4.71 & 4.64 & 4.28 & 4.34 & 9  & & 1567 (1472--1660) & 3  & 2 \\
babyshark            & 4.71   & 4.64   & 4.33   & 4.55   & 6  & & 1505 (1419--1592) & 4  & -- \\
Phd.ICU              & 4.66   & 4.60   & 4.24   & 4.39   & 10  & & 1494 (1403--1583) & 5  & -- \\
Softfoundry Facepro  & 4.62   & 4.57   & 4.43   & 4.59   & 7  & & 1479 (1400--1562) & 6  & -- \\
BlackShiba Labs      & 4.67   & 4.64   & 4.42   & 4.58   & 5  & & 1477 (1362--1577) & 7  & -- \\
Team 67              & 4.75   & 4.69   & 4.56   & 4.64   & 1  & & 1475 (1376--1582) & 8  & -- \\
teacher fat orange cat & 4.59 & 4.48   & 4.41   & 4.64   & 8 & & 1388 (1296--1476) & 9  & -- \\
Kuro                 & 4.74   & 4.67   & 4.53   & 4.59   & 2  & & 1380 (1269--1499) & 10 & -- \\
\midrule
All submissions (mean) & 4.31 & 4.19 & 3.77 & 3.86 & -- & & -- & -- & -- \\
\midrule
\multicolumn{10}{l}{\textit{Organizer baselines}} \\
Cascaded pipeline & 4.52 & 4.70 & 3.64 & 4.12 & -- & & -- & -- & -- \\
Commercial black-box generator (NotebookLM) & 4.53 & 4.48 & 4.00 & 4.26 & -- & & -- & -- & -- \\
Static retrieval framework & 2.12 & 2.47 & 3.53 & 4.01 & -- & & -- & -- & -- \\
\bottomrule
\end{tabular}
\end{table*}

 


\begin{table*}[t]
\centering
\small
\caption{
Crowd evaluation of the three organizer baselines under the Layer~2 protocol, each compared against the first-, second-, and fifth-ranked systems. Entries are wins across nine comparisons, and the final column indicates each baseline's placement.
}
\label{tab:baseline-human}
\begin{tabular}{lccc c}
\toprule
 & \multicolumn{3}{c}{Pairwise wins (out of 9)} & Inferred \\
\cmidrule(lr){2-4}
Baseline & vs.\ PhD.ICU (L2 \#5) & vs.\ ScaffoldAI (L2 \#2) & vs.\ tsunumon (L2 \#1) & Slot \\
\midrule
Cascaded Pipeline & 6/9 & 6/9 & 2/9 & 1--2 \\
Commercial Black-Box Generator (NotebookLM) & 8/9 & 8/9 & 4/9 & 1--2 \\
Static Retrieval Framework & 8/9 & 6/9 & 3/9 & 1--2 \\
\bottomrule
\end{tabular}
\end{table*}
 
\subsection{RQ1: Where Do Systems Succeed and Struggle?}
\label{sec:rq1}
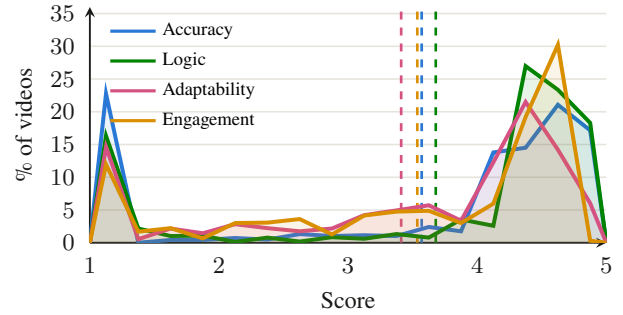
\begin{figure}[t]
    \centering

    \definecolor{acccol}{HTML}{2A78D6}
    \definecolor{logcol}{HTML}{008300}
    \definecolor{adacol}{HTML}{D55181}
    \definecolor{engcol}{HTML}{D98F00}
    \definecolor{ink}{HTML}{1A1A19}
    \definecolor{sec}{HTML}{575650}
    \definecolor{mut}{HTML}{8A887F}
    \definecolor{gridc}{HTML}{E7E4DD}

    \begin{tikzpicture}
    \begin{axis}[
      width=\columnwidth,
      height=4.7cm,
      xmin=1, xmax=5,
      ymin=0, ymax=36,
      xtick={1,2,3,4,5},
      ytick={0,5,10,15,20,25,30,35},
      xlabel={Score},
      ylabel={\% of videos},
      xlabel style={font=\small, color=ink},
      ylabel style={font=\small, color=ink},
      tick label style={font=\small, color=ink},
      axis line style={color=ink, line width=0.7pt},
      x tick style={draw=none},
      y tick style={draw=none},
      axis x line=bottom,
      axis y line=left,
      ymajorgrids=true,
      grid style={color=gridc, line width=0.6pt},
      clip mode=individual,
      legend style={
        at={(0.02,0.98)},
        anchor=north west,
        draw=none,
        fill=none,
        font=\scriptsize,
        legend cell align=left,
        row sep=1pt,
      },
      legend image post style={line width=1.2pt},
    ]

\addplot[acccol, dashed, line width=1.0pt, forget plot] coordinates {(3.570,0) (3.570,36)};
\addplot[logcol, dashed, line width=1.0pt, forget plot] coordinates {(3.679,0) (3.679,36)};
\addplot[adacol, dashed, line width=1.0pt, forget plot] coordinates {(3.411,0) (3.411,36)};
\addplot[engcol, dashed, line width=1.0pt, forget plot] coordinates {(3.536,0) (3.536,36)};
\addplot[acccol, line width=1.5pt, fill=acccol, fill opacity=0.09, draw opacity=1] coordinates {(1.000,0.0000) (1.125,22.7574) (1.375,0.0000) (1.625,0.4214) (1.875,0.4214) (2.125,0.7225) (2.375,0.4816) (2.625,1.3245) (2.875,1.0235) (3.125,1.1439) (3.375,1.0235) (3.625,2.4082) (3.875,1.7459) (4.125,13.7869) (4.375,14.5093) (4.625,21.0716) (4.875,17.1583) (5.000,0.0000)};
\addlegendentry{Accuracy}
\addplot[logcol, line width=1.5pt, fill=logcol, fill opacity=0.09, draw opacity=1] coordinates {(1.000,0.0000) (1.125,16.3155) (1.375,2.1674) (1.625,1.0235) (1.875,1.0837) (2.125,0.1204) (2.375,0.7827) (2.625,0.1806) (2.875,0.8429) (3.125,0.6020) (3.375,1.3245) (3.625,0.7827) (3.875,3.5521) (4.125,2.5888) (4.375,26.9717) (4.625,23.3594) (4.875,18.3022) (5.000,0.0000)};
\addlegendentry{Logic}
\addplot[adacol, line width=1.5pt, fill=adacol, fill opacity=0.09, draw opacity=1] coordinates {(1.000,0.0000) (1.125,14.6297) (1.375,0.5418) (1.625,2.1674) (1.875,1.4449) (2.125,2.8296) (2.375,2.2276) (2.625,1.7459) (2.875,2.1674) (3.125,4.2143) (3.375,4.9368) (3.625,5.7194) (3.875,3.3715) (4.125,12.2818) (4.375,21.4931) (4.625,14.2083) (4.875,6.0205) (5.000,0.0000)};
\addlegendentry{Adaptability}
\addplot[engcol, line width=1.5pt, fill=engcol, fill opacity=0.09, draw opacity=1] coordinates {(1.000,0.0000) (1.125,11.9807) (1.375,1.6857) (1.625,2.2276) (1.875,0.6623) (2.125,3.0102) (2.375,3.0704) (2.625,3.6123) (2.875,1.2643) (3.125,4.1541) (3.375,4.7562) (3.625,4.8766) (3.875,3.0102) (4.125,5.9603) (4.375,19.2053) (4.625,30.2228) (4.875,0.3010) (5.000,0.0000)};
\addlegendentry{Engagement}

    \end{axis}
    \end{tikzpicture}

    \caption{Distribution of videos across the four instructional quality dimensions. Solid lines represent score distributions, and dashed vertical lines indicate the respective means. The spike near a score of 1 in every dimension comes from teams that submitted off-topic or placeholder videos rather than competitive instructional content.}
    \label{fig:score_distribution}
\end{figure}

Table~\ref{tab:main} and Figure~\ref{fig:score_distribution} give the dimension-level answer.
The two content dimensions, Accuracy and Logic, score higher than the two delivery dimensions, Engagement and Adaptability, and the pattern holds for most individual systems.
Accuracy and Logic concentrate near the top of the scale, while Adaptability and Engagement spread into the middle and below, with Adaptability lowest overall.

To find the causes, we group the deductions in Table~\ref{tab:deductions} into four analytical categories.
These categories describe types of deficiency and are separate from the four evaluation dimensions.
Of the 6{,}699 deduction flags, visual delivery draws the largest share (39\%), led by ineffective visual representation, which draws a deduction in a third of the videos.
Learner adaptation is at 27\%, mostly due to missing scaffolding, jargon overload, and prerequisite gaps, so systems struggle to match explanations to the learner's assumed background.
Content problems are rarer (16\%) and mostly concern coverage and depth rather than correctness, with narration issues adding 8\%.
Explanation errors, counted separately by four count-based metrics, make up the remaining 10\% of flags, and about one video in six still contains a critical factual error, so high content scores coexist with occasional explanation mistakes.
\begin{figure}[t]
    \centering

    \definecolor{floorcol}{HTML}{8A887F}
    \definecolor{aicol}{HTML}{2A78D6}
    \definecolor{humancol}{HTML}{008300}
    \definecolor{ink}{HTML}{1A1A19}
    \definecolor{sec}{HTML}{575650}
    \definecolor{mut}{HTML}{8A887F}
    \definecolor{gridc}{HTML}{E7E4DD}
 
    \begin{tikzpicture}
    \begin{axis}[
      width=\columnwidth,
      height=4.5cm,
      ybar,
      bar width=24pt,
      xmin=0.4, xmax=3.6,
      ymin=0, ymax=86,
      xtick={1,2,3},
      xticklabels={
        {Persona-independent\\baseline},
        {Experimental},
        {Human-video}},
      xticklabel style={font=\scriptsize, align=center, color=black, yshift=-2pt},
      ytick={0,20,40,60,80},
      ylabel={Persona-identification accuracy (\%)},
      ylabel style={font=\scriptsize, color=black},
      tick label style={font=\small, color=black},
      axis line style={color=black},
      x tick style={draw=none},
      y tick style={draw=none},
      axis x line=bottom,
      axis y line=left,
      ymajorgrids=true,
      grid style={color=gridc, line width=0.6pt},
      clip mode=individual,
    ]
 
    \addplot+[ybar, bar shift=0pt, draw=floorcol, fill=floorcol!22,
      error bars/.cd, y dir=both, y explicit,
      error bar style={line width=0.9pt, color=ink}]
      coordinates {(1,30.8) += (0,13.5) -= (0,10.9)};
    \addplot+[ybar, bar shift=0pt, draw=aicol, fill=aicol!22,
      error bars/.cd, y dir=both, y explicit,
      error bar style={line width=0.9pt, color=ink}]
      coordinates {(2,51.7) += (0,8.7) -= (0,8.9)};
    \addplot+[ybar, bar shift=0pt, draw=humancol, fill=humancol!22,
      error bars/.cd, y dir=both, y explicit,
      error bar style={line width=0.9pt, color=ink}]
      coordinates {(3,61.9) += (0,13.1) -= (0,15.1)};
 
    \draw[dashed, sec, line width=0.8pt] (axis cs:0.45,33.3) -- (axis cs:3.55,33.3);
 
    \node[font=\scriptsize, color=ink] at (axis cs:1,48) {30.8\%};
    \node[font=\scriptsize, color=ink] at (axis cs:2,64) {51.7\%};
    \node[font=\scriptsize, color=ink] at (axis cs:3,80) {61.9\%};
 
    \end{axis}
    \end{tikzpicture}

    \caption{%
    Persona-identification accuracy by group, the rate at which raters pick the intended learner of a video from three candidates. Bars show per-group accuracy with Wilson 95\% confidence intervals, and the dashed line marks the 33.3\% chance rate.}
    \label{fig:persona_recovery}
\end{figure}
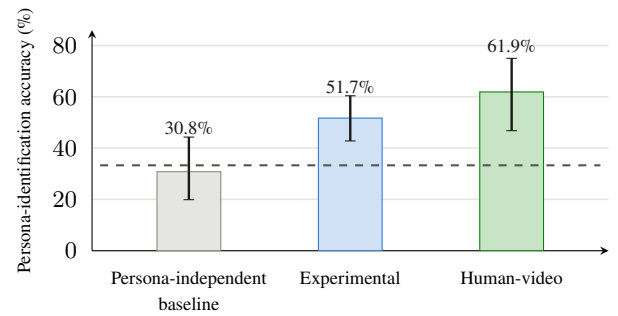

In short, current systems can cover and organize content well but still deliver and adapt it poorly.

\begin{table}[t]
\centering
\small
\caption{
Distribution of Layer~1 score deductions across 1,612 preliminary-round videos. Category percentages represent the share of the 6,699 total negative flags. Metric percentages indicate the proportion of videos triggering that specific metric (non-mutually exclusive).
}
\label{tab:deductions}
\setlength{\tabcolsep}{5pt}
\begin{tabular}{p{0.24\linewidth}p{0.50\linewidth}r}
\toprule
Category & Metric & (\%) \\
\midrule

\multicolumn{3}{l}{\textbf{Severity-scored metrics}}\\
\multirow{8}{*}{\textbf{Visual} \textit{(39\%)}}
& ineffective visual representation & 33\% \\
& AI-generated fatigue & 27\% \\
& narration--visual disconnect & 23\% \\
& visual signaling & 24\% \\
& visual alignment issue & 18\% \\
& decorative eye-candy & 11\% \\
& visual accessibility & 10\% \\
& visual clutter & 9\% \\

\addlinespace

\multirow{5}{*}{\textbf{Learner}\textit{(27\%)}}
& missing scaffolding & 29\% \\
& jargon overload & 23\% \\
& prerequisite gap & 22\% \\
& pacing mismatch & 18\% \\
& scaffolding failure & 16\% \\

\addlinespace

\multirow{4}{*}{\textbf{Content}\textit{(16\%)}}
& title--depth mismatch & 23\% \\
& content completeness & 21\% \\
& explanatory depth & 20\% \\
& pure calculation bias & $<1$\% \\

\addlinespace

\multirow{1}{*}{\textbf{Narration}\textit{(8\%)}}
& monotone / robotic audio & 33\% \\

\midrule
\multicolumn{3}{l}{\textbf{Count-based factual-error metrics}}\\

\multirow{4}{*}{\shortstack{\textbf{Explanation}\\\textbf{errors}\\\textit{(10\%)}}}
& critical fact error & 17\% \\
& causal inconsistency & 12\% \\
& minor slip & 9\% \\
& information overload & 4\% \\

\bottomrule
\end{tabular}
\end{table}

\subsection{RQ2: Do Systems Adapt to the Persona?}
\label{sec:rq2}
 
We next ask whether current systems truly adapt to the specified learner, since the aggregate Layer~1 scores cannot separate a system that adapts poorly from one that does not adapt at all.
We probe this with a human study that asks whether a human can recognize the learner for which a video was made from a video.
 
\paragraph{Persona identification study.}
A rater watches one instructional video and then chooses a persona that the video was made for among three candidate learner personas that differ in grade level and prior knowledge, so the chance baseline is one in three.
The three options are matched in length and format and shown in random order.
The study spans three groups of videos over the same eight topics, two from each subject.
The \emph{experimental} group is the output of the ten shortlisted systems, two videos per system.
The \emph{persona-independent baseline} is the output of the static retrieval framework, which selects a video by content relevance without adapting it to the learner, serving as a lower anchor.
The \emph{human-video} group is eight Creative Commons instructional videos with clearly identifiable audiences, one per item, whose best-fit persona was assigned by two researchers and reconciled by discussion, serving as a topline.
 
\paragraph{Results.}
In total, 69 Prolific raters contributed 214 ratings, summarized in Figure~\ref{fig:persona_recovery}.
For the persona-independent baseline, the intended learner is identified only at the chance level.
For the experimental group, it is identified well above this baseline, so their adaptation is perceptible to raters.
Because the ratings are clustered by video and by rater and thus not independent, we confirm the gap using a mixed-effects logistic regression with random intercepts for both, and the gap remains (full coefficients in Appendix~\ref{app:persona-model}).
The learner persona in the human-video group is identified more often than in the experimental group, but the difference is not statistically significant.
In response to RQ2, systems do adapt to the specified learner in ways that people can perceive.
Their low Learner Adaptability score in RQ1 therefore does not mean they fail to adapt, only that they do not adapt well.
 
\subsection{RQ3: Does the LLM-Judge Align with Human Judgment?}
\label{sec:rq3}

We next evaluate how well the LLM-judge aligns with human judgment, analyzing the 320 preliminary-round videos and the 48 final videos.
The short answer is that the judge separates clearly weak submissions from the rest but ranks the strongest systems poorly.

\paragraph{The LLM-Judge Ranks the Strongest Systems Poorly.}
Table~\ref{tab:main} shows the ten shortlisted systems all scoring near the ceiling of 5, which leaves the judge little room to tell them apart.
As a result, its ranking of these ten has almost no agreement with the Layer~2 crowd ranking (Spearman $\rho=-0.17$).
\texttt{tsunumon}, the eventual winner, ranks only fourth under the judge, while the judge's top two, \texttt{Team~67} and \texttt{Kuro}, finish eighth and tenth with the crowd.
This mismatch is not measurement noise on either side.
On the judge's side, scores are stable across repeated runs, as the rescoring check in the appendix shows.
On the crowd's side, \texttt{tsunumon}'s 90\% Elo interval overlaps neither \texttt{Team~67}'s nor \texttt{Kuro}'s.

\paragraph{Prevalence vs. Salience in Evaluation Criteria.}
To see why the judge and the crowd disagree, we compare what each emphasizes.
For the 320 preliminary videos, we group the judge's metrics and the crowd's written rationales, available for 145 of the 246 comparisons, into the four categories of Section~\ref{sec:rq1}.
The judge is summarized by prevalence, the share of videos on which a criterion is recorded, favorably or not.
The crowd is summarized by salience, the share of rationales that cite a criterion as the reason for a preference.
Table~\ref{tab:attention} lists the top ten of each.
The two broadly agree on what matters, since learner adaptation and content coverage fill eight of the crowd's ten reasons and six of the judge's ten criteria.

The rankings still differ because the judge gives the top systems almost the same scores.
Criteria the judge records heavily but the crowd seldom cites apply to nearly every video or to almost none, so they separate nothing.
Pure-calculation bias, for example, is recorded on 99.7\% of videos and almost always favorably.
Even on the criteria both sides value, the top systems' scores are nearly identical.
Ranking on such scores amounts to ranking on tiny, essentially random differences, while the crowd's pairwise choices turn on the small differences their rationales actually cite.
The judge and the human layers are thus complementary, one screening every video and the other comparing the strongest few.

\begin{table*}[t]
\centering
\caption{What the Layer 1 judge and the Layer 2 crowd each emphasize in the preliminary round. Left: the ten Layer 1 metrics that register most frequently, as the percentage of the 320 preliminary-round videos submitted by the ten shortlisted systems; a registration can be favorable, so high prevalence reflects how often the judge records a criterion, not how often videos fail it. Right: the ten reasons cited most frequently in the 145 written pairwise-comparison rationales, as the percentage of rationales that mention each. The two halves are not on a common scale and are read down, not across.}
\label{tab:attention}
\begin{tabular}{@{}l S[table-format=2.1] @{\hspace{4em}} l S[table-format=2.1]@{}}
\toprule
Metric & {\% of 320 videos} & Reason & {\% of 145 rationales} \\
\midrule
pure-calculation bias (content)            & 99.7 & visuals (visual)          & 33.8 \\
visual signaling (visual)                  & 97.5 & coverage (content)        & 29.7 \\
visual alignment issue (visual)            & 96.6 & examples (adaptation)     & 26.2 \\
scaffolding failure (adaptation)           & 88.1 & level (adaptation)        & 25.5 \\
narration-visual disconnect (visual)       & 82.2 & structure (content)       & 24.8 \\
ineffective visual representation (visual) & 81.6 & scaffolding (adaptation)  & 24.1 \\
missing scaffolding (adaptation)           & 67.2 & depth (content)           & 24.1 \\
explanatory depth (content)                & 45.6 & voice (narration)         & 20.7 \\
pacing mismatch (adaptation)               & 22.5 & concise (adaptation)      & 17.2 \\
content completeness (content)             &  9.7 & interactivity (adaptation)& 11.0 \\
\bottomrule
\end{tabular}
\end{table*}

\section{Discussion}

\subsection{A Qualitative Case Study of a Finalist Video}
\label{sec:case-study}

\begin{figure*}[t]
\centering
\includegraphics[width=0.48\textwidth]{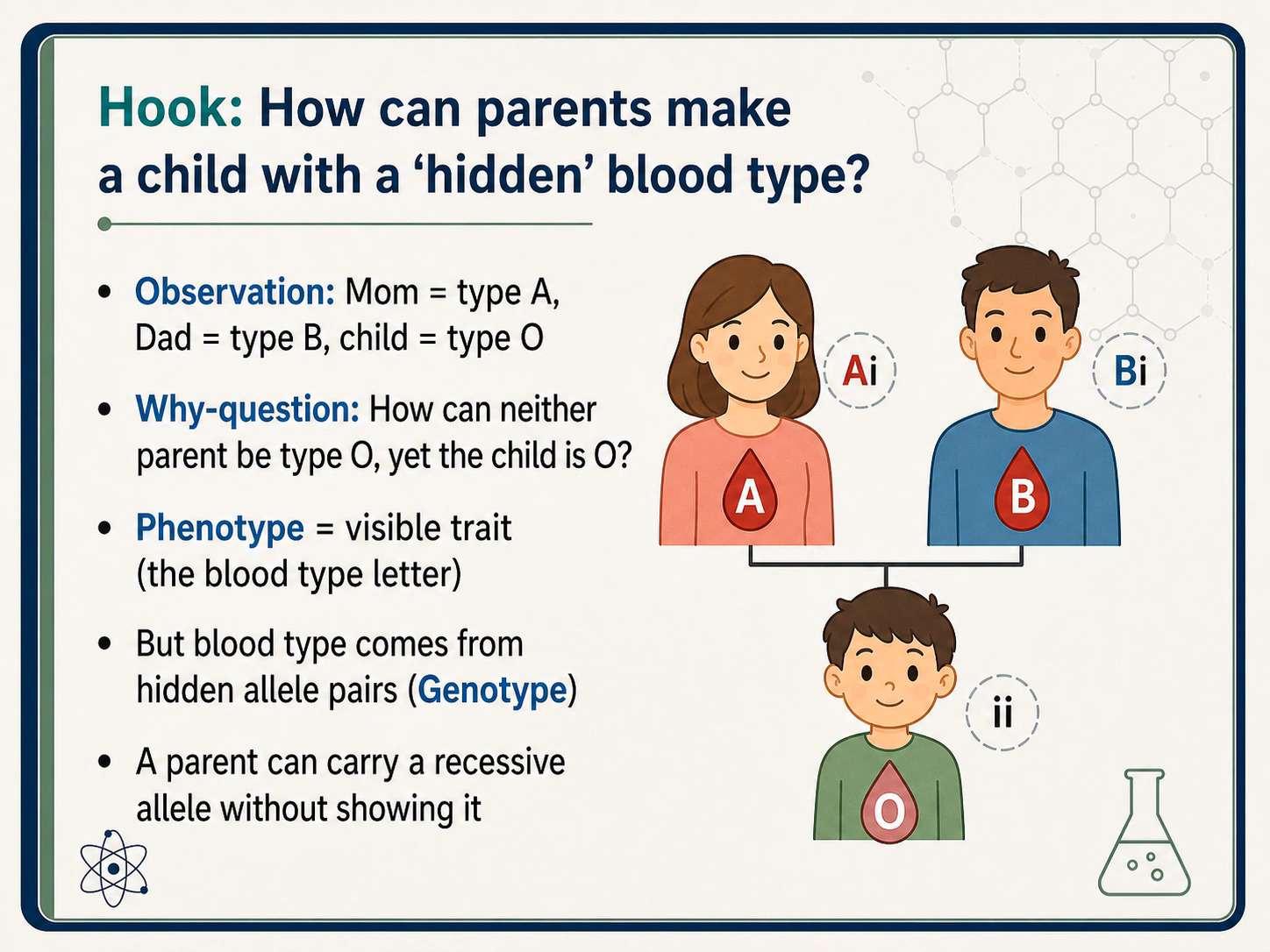}
\hfill
\includegraphics[width=0.48\textwidth]{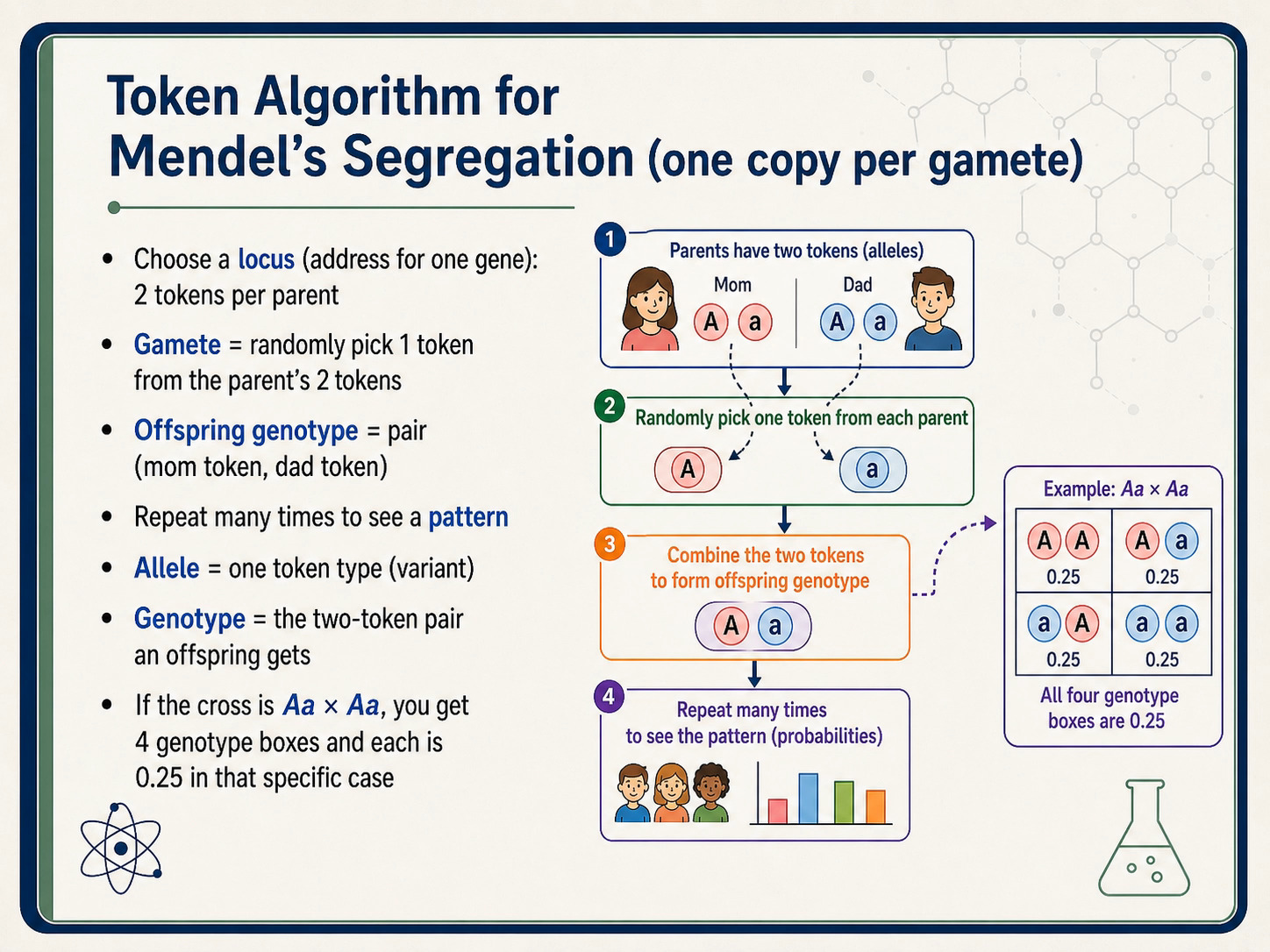}
\caption{Two slides from a finalist instructional video on genetics. Left: the hook slide introduces a blood-type puzzle. Right: the token-algorithm slide explains Mendelian segregation.}
\label{fig:ai-teach-limit-genetics}
\end{figure*}

To complement the preceding quantitative analyses, we examine two slides from one of the three finalist videos.
Figure~\ref{fig:ai-teach-limit-genetics} shows a final-round instructional video on Mendelian genetics.
The hook slide poses a blood-type puzzle, while the token-algorithm slide explains Mendel's law of segregation by modeling each parent as contributing one allele.
Although illustrative rather than representative, this case shows how a high-ranking video can remain limited across the four evaluation dimensions introduced in Section \ref{sec:llm-judge-sreening}.

\paragraph{Content Accuracy.}
The two slides are broadly accurate and use basic terminology appropriately.
The hook slide distinguishes phenotype from genotype, while the token-algorithm slide correctly shows that an offspring receives one allele from each parent.
Several statements, nevertheless, lack precision.
The title ``Mendel's Segregation'' is less precise than ``Mendel's law of segregation,'' and ``Gamete = randomly pick 1 token'' describes an action rather than the resulting reproductive cell.
More importantly, the phrase ``4 genotype boxes'' can be read as four distinct genotypes.
An $Aa \times Aa$ cross has four equally likely gamete-pair outcomes but only three genotype categories because two cells represent $Aa$.
Their probabilities are $AA=0.25$, $Aa=0.50$, and $aa=0.25$.
The wording therefore conflates Punnett-square cells with genotype categories.

The generated visuals introduce further ambiguity.
At Step 4 of the token-algorithm slide, children with different skin tones appear even though skin tone is never defined as the modeled trait.
This may suggest that the $A/a$ alleles directly encode racial appearance, and the dark-skinned child appears without any explanation for that visual variation.
In the Punnett square inset, colors also cease to consistently mark maternal and paternal origin.
Keeping one red maternal token and one blue paternal token in every cell would preserve the one-allele-per-parent rule.

\paragraph{Pedagogical Logic.}
The video follows a clear progression from a concrete puzzle to a procedural explanation supported by text, diagrams, and a probability grid.
However, the hook slide uses ABO blood type, whereas the token-algorithm slide shifts to a generic $Aa \times Aa$ cross without explaining the abstraction.
The token-algorithm slide also introduces ``locus'' without using it later and repeats similar content across the text list, flowchart, and Punnett square.
Integrating these representations and preserving consistent conventions would reduce attention switching without removing content.

\paragraph{Learner Adaptability.}
The learner persona specified a seventh-grade learner with a 25-minute focus time, basic knowledge of reproduction and probability, and no prior formal genetics knowledge.
The requested objective was to analyze how alleles are distributed to offspring in classical genetics.
The hook slide fits this persona by using a familiar human blood-type example, while the token-algorithm slide directly models allele distribution.
However, several technical terms are introduced in rapid succession, and the explanation skips from random parental selection to four outcomes labeled 0.25.
For a learner with no formal genetics training and only basic probability knowledge, the video should explicitly derive each outcome from two independent choices, $0.5 \times 0.5 = 0.25$.

\paragraph{Engagement and Multimodal Presentation.}
The question-based hook slide, familiar example, orderly layout, and large illustrations make the video visually engaging.
However, visual polish does not ensure effective multimodal teaching.
In our review, some technical terms and formulas in the text-to-speech narration were difficult to discern, while the delivery used limited prosodic variation and little slowing during the probability explanation.
The token-algorithm slide is also dense, combining terminology, a flowchart, a Punnett square, and a probability summary.

The video also provides no cursor or pointer to guide attention.

This case illustrates, at the level of a single finalist video, the pattern quantified in Table \ref{tab:main} and Section \ref{sec:rq1}: strong content scores coexisting with weaknesses in multimodal signaling and learner adaptation. We use it to make those aggregate deductions concrete, not as independent evidence of a capability ceiling.

\subsection{Feedback from Participants}
\label{sec:participant-feedback}
As detailed in Appendix~G, feedback from 15 teams indicated that while the end-to-end challenge design was highly valued, primary engineering frictions stemmed from multi-modal synchronization under strict latency constraints and ambiguous LLM-judge feedback. These self-reported concerns regarding evaluator noise align directly with the near-zero judge-human correlation found in Section~7.4. Ultimately, this highlights the critical need for transparent, inspectable evaluation trails in future benchmarks.

\section{Limitations}
\label{sec:limitations}

The first Teaching Monster Challenge was limited to AP-aligned STEM subjects at the secondary-school level, including Physics, Biology, Computer Science, and Mathematics, and all instructional videos were generated in English. The findings may not generalize to other subject areas or non-English educational settings. Future work can extend the benchmark to a broader range of subjects and languages.

Human evaluation in this study also reflects a specific educational context. The final expert panel consisted mainly of secondary-school teachers, school principals, and university professors with experience in Taiwan's education system. Their judgments reflect the instructional practices and evaluation traditions of Taiwanese secondary education, rather than a universal standard across educational systems. Future work could include more internationally diverse expert panels.

This benchmark evaluates single-shot instructional video generation rather than interactive tutoring. Systems generate a complete instructional episode from a course requirement and learner persona, but do not adapt to later learner questions, errors, or confusion. The challenge measures the ability to turn learning needs into structured instructional content, rather than continuous diagnosis, adaptation, and individualized feedback. Future work could incorporate multi-turn interaction and learner responses.

\section{Conclusion}
We introduced the Teaching Monster Challenge, the first instructional video generation benchmark to make learner adaptation, the transformation at the core of PCK, an explicit evaluation criterion.
The first edition shows that current agents have the content knowledge teaching rests on.
They mostly get the material right and organize it into coherent lessons, yet they fall short in presenting it clearly and adapting it to the learner.
People can recognize whom a lesson was made for, so the adaptation PCK requires is present, only not yet done well.

Judging the systems proved a challenge in itself.
The judge screens out clearly weak submissions, but the best systems all score near the top of its scale, so its ranking of them no longer matches human preference.
In this edition, the judge decided who advanced, and only humans could decide who won.
Progress on this task therefore has two fronts: better teaching systems and better judges of teaching.
We release the items, rubric, scoring traces, and human judgments as one testbed for both.
Later editions will move past AP-level STEM and single-shot generation toward interactive tutoring, where teaching is hardest and matters most.

\section*{Acknowledgments}
We thank Shou-De Lin for his advice and discussions throughout this work.
We also thank the members of the expert panel who designed the test items and served as judges for the Layer~3 evaluation: Tsung-Hau Jen, Chi-Ta Chia, Pei-Chun Liao, Chia-Hui Cheng, Kuo Yuan Lin, J~Will James, Shihkuan Hsu, and Hsiou-Huai Wang.
This work was supported by the Ministry of Education (MOE) of Taiwan under the project Taiwan Centers of Excellence in Artificial Intelligence, through the NTU Artificial Intelligence Center of Research Excellence (NTU AI-CoRE).

\bibliography{references}

\appendix
\section{Origin of the Name Teaching Monster}
\label{app:appendix.orign.tm}
The name of the challenge pays tribute to two works of popular culture.
The first is the reality show Chinese Monster, produced by the YouTuber Ku's Dream, whose popularity started a trend of naming competitions after the word monster.
In that context, the word sheds its usual negative connotation and instead denotes a top performer who breaks conventions in a given field, and the challenge adopts the word in the same spirit, in the hope that AI agents can become teaching monsters that break the frame of conventional instruction.
The second is the anime Assassination Classroom, whose teacher character, Koro-sensei, is depicted as a monster capable of moving at Mach 20, an ability he uses to prepare tailored materials and give one-on-one guidance to every student in the class at the same time.
This vision of fully individualized instruction for every learner was once imaginable only in fiction, and the challenge asks whether AI agents can begin to provide personalized instruction of that kind for every student.

\section{LLM-Judge Scoring Protocol}

\subsection{Scoring Mechanism}
\label{sec:l1_scoring}
The scoring procedure conducted by the LLM-judge is divided among three agents working in collaboration.
First, the perceiving agent watches the original video, extracts a time-stamped content map, and conducts a multimodal audit, but does not itself assign scores.
Second, the grading agent produces the Content Accuracy and Pedagogical Logic scores based only on the content map and the multimodal audit results of the perceiving agent.
Third, the persona judging agent, while referring to the structured output of the perceiving agent, watches the original video again and scores the two more subjective dimensions, Learner Adaptability and Engagement and Multimodal Presentation.

In the scoring mechanism for the LLM-judge, each dimension has an independent score that starts from a baseline of \(4.0\).
The score is then adjusted according to the performance of each indicator within the dimension, through a penalty, a bonus when expectations are exceeded, or no effect on the score.
For penalties, indicators scored by severity, such as the degree of agreement between title and content, are deducted in three graded levels according to the severity of the violation of each indicator.
Indicators scored by count, such as factual errors, are deducted according to the number of occurrences multiplied by the deduction per occurrence, although the deduction for a single count-based penalty is capped at \(2.0\) points.
All penalties have different deduction values for each indicator, which are detailed below.
The bonus, in turn, is computed according to the number of bonus items obtained in each dimension, adding to the score the number of bonus items obtained divided by the total number of bonus items available in that dimension, which means that the upper bound of the bonus is \(1.0\).
The final score is clipped to \([1.0, 5.0]\).

\subsection{Rubric Items}

Detailed rating rubrics are shown in Tables \ref{tab:Metric1}, \ref{tab:Metric2}, \ref{tab:Metric3}, and \ref{tab:Metric4}.
Note that Content Accuracy and Pedagogical Logic share four sub-items: Scaffolding Failure, Pure Calculation Bias, Content Completeness, and Explanatory Depth.
This is because a deficiency in any one of these four indicators will inevitably harm both Content Accuracy and Pedagogical Logic.

\subsection{LLM-Judge Consistency}
\label{app:judge_consistency}

To assess the stability of the LLM-judge, we repeated the evaluation of a subset of videos and compared the resulting scores. 
We define a \emph{(system, topic) cell} as one evaluation instance of a given system on a given topic. Among 3 repeated evaluations of the same cell, the median within-cell standard deviation of the overall score is 0.14 on the 1.0--5.0 scoring scale, indicating that repeated judgments are generally stable. 
Averaging multiple evaluations therefore provides a sufficiently robust signal for admit-or-reject screening.


\begin{table*}[hbt]
\centering
\small
\setlength{\tabcolsep}{1mm}
\caption{Content Accuracy scoring metrics provided to LLM-judge}
\label{tab:Metric1}
\begin{tabularx}{\textwidth}{@{} p{1.9cm} Y r @{}}
\toprule
\textbf{Metric} & \textbf{Rating description} & \textbf{Value} \\
\midrule
\multirow{4}{1.9cm}{Scaffolding Failure: Formula / Logic Leap (shared)}
 & Every major formula is motivated and derivations are step-complete; scaffolding clearly stronger than typical & bonus \\
 & Skips the ``why'' for 1 key concept, OR 1--2 unjustified derivation jumps & $-0.5$ \\
 & Multiple key formulas lack rationale, OR multiple critical steps skipped & $-1.0$ \\
 & Entire video presents formulas as facts with zero scaffolding, OR derivations riddled with leaps & $-1.5$ \\
\midrule
\multirow{4}{1.9cm}{Pure Calculation Bias (shared)}
 & Calculations subordinate to a visible conceptual spine; concept-first feel & bonus \\
 & Theory exists but brief ($\sim$60\% calculation) & $-0.3$ \\
 & Heavily skewed ($\sim$75\% calculation) & $-0.6$ \\
 & Almost pure calculation ($>$85\%) & $-1.0$\\
\midrule
\multirow{4}{1.9cm}{Content Completeness (shared)}
 & Content notably richer than the title implies; goes beyond stated scope & bonus \\
 & 1 minor supporting concept missing, or content slightly thinner than implied & $-0.5$ \\
 & 1 core concept absent, OR multiple supporting concepts missing, OR video too short & $-1.0$ \\
 & Multiple core concepts absent; fails stated scope; OR near-empty & $-1.5$ \\
\midrule
\multirow{4}{1.9cm}{Explanatory Depth (shared)}
 & Depth clearly above typical and sustained; critical steps not skipped, assumptions explicit & bonus \\
 & Some topics shallower than the title implies; key steps occasionally rushed; a few only ``mentioned'' & $-0.5$ \\
 & Most topics ``mentioned'' without explanation; OR title implies derivation but content heavily abbreviated & $-1.5$ \\
 & Title promises concept/derivation but delivery entirely procedural; OR topics merely listed & $-2.0$ \\
\midrule
\multirow{3}{1.9cm}{Title--Depth Mismatch}
 & Minor overpromising (small scope or depth difference) & $-0.5$ \\
 & Clear mismatch (e.g.\ ``Derivation'' video shows only a formula) & $-2.0$ \\
 & Fundamental mismatch (different topic, or concept promised but video is plug-and-chug) & $-4.0$ \\
\midrule
\multirow{2}{1.9cm}{Visual Alignment Issue}
 & Visuals track the spoken argument in real time; anticipate/repair confusion & bonus \\
& Occasional misalignment & $-0.2$ \\
 & Decorative images during technical content; slides lag & $-0.5$ \\
 & Visuals actively contradict or distract & $-1.0$ \\
\midrule
Critical Fact Error (count) & Each uncorrected scientific/mathematical error & $-0.3$ per counting \\
\midrule
Minor Slip (count) & Each persistent minor error (typo, dropped sign, imprecise term) & $-0.2$ per counting \\
\bottomrule
\end{tabularx}
\end{table*}

\begin{table*}[ht]
\centering
\small
\setlength{\tabcolsep}{1mm}
\caption{Pedagogical Logic scoring metrics provided to LLM-judge}
\label{tab:Metric2}
\begin{tabularx}{\textwidth}{@{} p{1.9cm} Y r @{}}
\toprule
\textbf{Metric} & \textbf{Rating description} & \textbf{Value} \\
\midrule
\multirow{4}{1.9cm}{Scaffolding Failure: Formula / Logic Leap (shared)}
 & Every major formula is motivated and derivations are step-complete; scaffolding clearly stronger than typical & bonus \\
 & Skips the ``why'' for 1 key concept, OR 1--2 unjustified derivation jumps & $-0.5$ \\
 & Multiple key formulas lack rationale, OR multiple critical steps skipped & $-1.0$ \\
 & Entire video presents formulas as facts with zero scaffolding, OR derivations riddled with leaps & $-1.5$ \\
\midrule
\multirow{4}{1.9cm}{Pure Calculation Bias (shared)}
 & Calculations subordinate to a visible conceptual spine; concept-first feel & bonus \\
 & Theory exists but brief ($\sim$60\% calculation) & $-0.3$ \\
 & Heavily skewed ($\sim$75\% calculation) & $-0.6$ \\
 & Almost pure calculation ($>$85\%) & $-1.0$ \\
\midrule
\multirow{4}{1.9cm}{Content Completeness (shared)}
 & Content notably richer than the title implies; goes beyond stated scope & bonus \\
 & 1 minor supporting concept missing, or content slightly thinner than implied & $-0.5$ \\
 & 1 core concept absent, OR multiple supporting concepts missing, OR video too short & $-1.0$ \\
 & Multiple core concepts absent; fails stated scope; OR near-empty & $-1.5$ \\
\midrule
\multirow{4}{1.9cm}{Explanatory Depth (shared)}
 & Depth clearly above typical and sustained; critical steps not skipped, assumptions explicit & bonus \\
 & Some topics shallower than the title implies; key steps occasionally rushed; a few only ``mentioned'' & $-0.5$ \\
 & Most topics ``mentioned'' without explanation; OR title implies derivation but content heavily abbreviated & $-1.5$ \\
 & Title promises concept/derivation but delivery entirely procedural; OR topics merely listed & $-2.0$ \\
\midrule
Causal Inconsistency (count) & Each conclusion stated without supporting logic/evidence & $-0.4$ per counting \\
\midrule
Information Overload (count) & Each segment cramming too much without clear transitions & $-0.2$ per counting \\
\bottomrule
\end{tabularx}
\end{table*}

\begin{table*}[ht]
\centering
\small
\setlength{\tabcolsep}{1mm}
\caption{Learner Adaptability scoring metrics provided to LLM-judge}
\label{tab:Metric3}
\begin{tabularx}{\textwidth}{@{} p{1.9cm} Y r @{}}
\toprule
\textbf{Metric} & \textbf{Rating description} & \textbf{Value} \\
\midrule
\multirow{3}{1.9cm}{Jargon Overload}
 & 1 slide with undefined terms & $-0.3$ \\
 & 2 slides & $-0.6$ \\
 & 3+ or video-wide style & $-1.0$ \\
\midrule
\multirow{4}{1.9cm}{Prerequisite Gap}
 & Proactively bridges background gaps beyond what the persona's knowledge requires & bonus \\
 & Speed bump: a few gaps, patchable while watching & $-0.3$ \\
 & Stop \& fix: must pause/rewind/Google repeatedly & $-0.6$ \\
 & Roadblock: core content incomprehensible, would give up & $-1.0$ \\
\midrule
\multirow{4}{1.9cm}{Pacing Mismatch}
 & Ideal rhythm for this persona's focus time and grade-level processing speed & bonus \\
 & Occasionally too fast/slow; still finishes & $-0.3$ \\
 & Often mismatched; frequently lost or time wasted & $-0.6$ \\
 & Unwatchable at this pace for this persona & $-1.0$ \\
\midrule
\multirow{3}{1.9cm}{Visual Accessibility}
 & 1 slide with clear legibility problem & $-0.3$ \\
 & 2 slides & $-0.6$ \\
 & 3+ or default style throughout & $-1.0$ \\
\midrule
\multirow{4}{1.9cm}{Missing Scaffolding}
 & Scaffolding goes beyond the persona's minimum needs, proactively anticipating confusion & bonus \\
 & 1 stretch feels under-explained & $-0.3$ \\
 & Major stretches missing preferred support & $-0.6$ \\
 & Opposite of need for long stretches; learning fails & $-1.0$ \\
\midrule
\multirow{4}{1.9cm}{Ineffective Visual Representation}
 & Visuals exceptionally illuminate concepts & bonus \\
 & 1 slide/beat missing a needed visual aid & $-0.3$ \\
 & 2 slides/beats; mental simulation tiring & $-0.6$ \\
 & 3+ or nearly all conceptual parts lack visuals & $-1.0$ \\
\bottomrule
\end{tabularx}
\end{table*}

\begin{table*}[ht]
\centering
\small
\setlength{\tabcolsep}{1mm}
\caption{Engagement and Multimodal Presentation scoring metrics provided to LLM-judge}
\label{tab:Metric4}
\begin{tabularx}{\textwidth}{@{} p{1.9cm} Y r @{}}
\toprule
\textbf{Metric} & \textbf{Rating description} & \textbf{Value} \\
\midrule
\multirow{4}{1.9cm}{Monotone / Robotic Audio}
 & Energizing delivery with clear vocal variety & bonus \\
 & Somewhat flat; mild urge to tune out & $-0.4$ \\
 & Drone-like / TTS chunking breaks immersion; rewind needed & $-0.8$ \\
 & Would stop or skip due to voice & $-1.2$ \\
\midrule
\multirow{3}{1.9cm}{AI Generated Fatigue}
 & 1 slide feels cheap/samey/uncanny & $-0.3$ \\
 & 2 instances & $-0.6$ \\
 & 3+ or dominant AI art style end-to-end & $-1.0$ \\
\midrule
\multirow{3}{1.9cm}{Visual Clutter}
 & 1 slide hard to scan while listening & $-0.3$ \\
 & 2 slides & $-0.6$ \\
 & 3+ or clutter is the norm & $-1.0$ \\
\midrule
\multirow{4}{1.9cm}{Narration / Visual Disconnect}
 & Visuals and narration tightly synchronized & bonus \\
 & 1 noticeable mismatch; meaning recoverable & $-0.3$ \\
 & 2 mismatches or 1 long misleading stretch & $-0.6$ \\
 & 3+ or visuals repeatedly contradict narration & $-1.0$ \\
\midrule
\multirow{3}{1.9cm}{Decorative Eye-Candy}
 & 1 segment of flashy but empty spectacle & $-0.3$ \\
 & 2 segments & $-0.6$ \\
 & 3+ or eye-candy replaces substance & $-1.0$ \\
\midrule
\multirow{3}{1.9cm}{Visual Signaling}
& Signaling is consistently timed and precise --- viewer always knows where to look, enhancing comprehension noticeably beyond baseline. & bonus \\
 & 1--2 moments viewer must hunt for the relevant element & $-0.3$ \\
 & Multiple segments narrate specifics with no visual guidance & $-0.6$ \\
 & Systematically no cueing throughout; cannot follow along & $-1.0$ \\
\bottomrule
\end{tabularx}
\end{table*}

\section{Organizer Baselines: Implementation Details}
\label{app:baselines}

\paragraph{Cascaded Pipeline.}
This baseline is a linear chain of self-contained, zero-shot services and models in which each stage consumes only the artifacts produced by earlier stages, with no joint optimization and no end-to-end feedback. It operated under the same 30-minute generation limit as the submitted systems.
It runs in three sequential phases.
First, in the planning phase, an outline module expands the \texttt{course\_requirement} and \texttt{learner\_persona} into a structured storyboard (\texttt{gemini-3-flash-preview} for context analysis, then \texttt{gemini-3.1-pro-preview} with search grounding), and a wrapper stage converts the storyboard into per-slide content specifications and narration scripts (\texttt{gemini-3-flash-preview}).
Second, in the asset generation phase, slides and narration are produced concurrently: each slide passes through a design-and-review loop in which an LLM designer proposes a layout, a deterministic renderer realizes it, and a vision-LLM reviewer requests fixes (\texttt{gemini-3.1-pro-preview}, inline images by \texttt{gemini-2.5-flash-image}), while narration is synthesized by an open-weight TTS model (\texttt{Qwen/Qwen3-TTS-12Hz-1.7B-CustomVoice}) and word-level timing is recovered by aligning the narration audio against a \texttt{faster-whisper} transcription.
Two open-weight vision-language models (\texttt{Qwen/Qwen3-VL-8B-Instruct} and \texttt{ByteDance-Seed/UI-TARS-1.5-7B}) then ground each narration segment to a coordinate on its slide, defining the path of an instructional pointer.
Finally, in the compilation phase, a deterministic rendering loop overlays the pointer, scheduling its movements at the aligned word timestamps, and concatenates the slide imagery and audio segment by segment, with each segment temporally bounded by its narration duration; this final phase is the only deterministic one, while the planning and slide-design stages sample from LLMs.
The storyboard fixes the slide count (bounded between three and ten), so the output video length follows from the narration rather than from an explicit persona-conditioned duration target.
Table~\ref{tab:cascade-config} summarizes the per-module configurations; full identifiers appear in the text above.

\begin{table}[t]
\centering
\resizebox{\columnwidth}{!}{%
\begin{tabular}{@{}lll@{}}
\toprule
Stage & Backbone & Key settings \\
\midrule
Context analysis & \texttt{gemini-3-flash-preview} & structured JSON \\
Storyboard & \texttt{gemini-3.1-pro-preview} & search grounding \\
Wrapper & \texttt{gemini-3-flash-preview} & structured JSON \\
Slide designer & \texttt{gemini-3.1-pro-preview} & temperature 0.7 \\
Slide reviewer & \texttt{gemini-3.1-pro-preview} & temp.\ 0.3; $\leq$5 rounds \\
Image generation & \texttt{gemini-2.5-flash-image} & image modality \\
Image check & \texttt{gemini-3.1-pro-preview} & temperature 0.2 \\
TTS & \texttt{Qwen3-TTS-1.7B} & single English voice \\
Alignment & \texttt{faster-whisper} (medium) & word timestamps \\
Pointer grouping & \texttt{Qwen3-VL-8B-Instruct} & greedy; 4-bit \\
Pointer grounding & \texttt{UI-TARS-1.5-7B} & greedy; 4-bit \\
Compilation & moviepy & 15 fps; H.264 \\
\bottomrule
\end{tabular}%
}
\caption{Module configurations of the cascaded pipeline baseline; the planning and wrapper calls use provider-default sampling.}
\label{tab:cascade-config}
\end{table}

\paragraph{Commercial Black-Box Generator.}
The black-box baseline drives NotebookLM's Video Overview feature programmatically rather than through the web interface, using the Explainer format with automatic style selection, and was accessed between May 1 and May 15, 2026.
For each item, a lightweight adapter derives a research query and a host-steering instruction from the two input fields; the service gathers its own source material through the query and generates under the instruction, which sets audience level, prerequisites, and target length according to the \texttt{learner\_persona}.
The first successfully generated video was used for every item, with retries only on technical failure and no quality-based selection among candidates. Outputs were downloaded as MP4 files and served to the evaluation platform.
Because the generation process is proprietary and closed, how these inputs shape the output cannot be verified, and the reported scores characterize the service as accessed at that time rather than a reproducible system.

\paragraph{Static Asset Retrieval Framework.}
The retrieval framework uses no generative language model at any stage. It retrieves an existing video in two steps, building a fixed corpus and then ranking that corpus for instructional relevance, and so marks the non-generative corner of the design space.
The corpus is assembled in advance from Creative Commons (CC) licensed instructional videos covering the target domain, so that retrieval runs against a fixed pool rather than the continually changing open web.
For each preliminary-phase task, candidate videos are gathered through CC-filtered YouTube searches, using queries derived automatically from the \texttt{course\_requirement}.
Each candidate's license is automatically verified from platform metadata, and filtering by duration and deduplication yield a frozen set of 257 videos.
Retrieval follows the standard cross-encoder reranking approach \cite{nogueira2019}: a cross-encoder (\texttt{cross-encoder/ms-marco-MiniLM-L-6-v2}) reads the query jointly with a candidate and scores whether the video actually teaches the requested content rather than merely mentioning its keywords, the relevance signal a bi-encoder alone misses.
The query is formed from the subject and \texttt{course\_requirement} alone; the \texttt{learner\_persona} is deliberately excluded so that content matching stays separate from personalization and does not inflate the persona-adaptation measured by the judge.
Because a full transcript far exceeds the cross-encoder's 512-token input window, a bi-encoder (\texttt{BAAI/bge-small-en-v1.5} \cite{10.1145/3626772.3657878}) first locates each video's most query-relevant transcript passage, and the cross-encoder reads that passage together with the title and description.
Two objective adjustments then refine the relevance score: one penalizes a subject mismatch, which prevents a keyword-similar video from another subject from displacing an on-subject one, and one penalizes a video whose length departs from the \texttt{learner\_persona}'s stated attention span.
The attention span is the only persona signal admitted, used as a duration constraint rather than for content matching; by design the persona influences only the admissible length, so the baseline probes how much of the task pure content retrieval can satisfy and provides a calibration point for the judge's Learner Adaptability dimension.
The cross-encoder's top-ranked video is returned as the answer.

\emph{Reproducibility.}
The cross-encoder reads each candidate's title, description, and most query-relevant transcript passage, truncated to its 512-token window.
Its relevance score is adjusted by an off-subject penalty of 3.0 and a duration penalty of $2.0 \times (1 - f)$, where $f$ equals 1 when the video runs between 0.4 and 1.5 times the attention span and decays to 0 at 3 times the span.

\section{Example Question Sets}
\label{app:example-question-sets}

This appendix shows two worked matched-pair examples drawn from the first edition. The Type A pair varies only the student persona to isolate the persona-conditioning effect. The Type B pair holds the topic fixed and raises the target cognitive level, so the learning objective is rewritten to demand higher-order skills.

\paragraph{Type A example: persona contrast.}
The two items below adapt the same lesson for an 8th grader (Item 1) and a 1st-year university student (Item 2).

\examplequestionbox
{typeAblue}
{typeAbackground}
{Item 1 (Type A: persona contrast)}
{%
\textbf{Topic:} Overfitting vs.\ Underfitting: Why the Best Model Is Not Always the Most Complex.\par
\textbf{Learning objective:} Deliver a clear lecture explaining the concepts of overfitting and underfitting, visually guiding students through the ``training/test error vs.\ model complexity'' curves and explicitly identifying the regions corresponding to each.\par
\textbf{Bloom-level:} \{Remember, Understand\}.\par
\textbf{Grade level:} 8th grade.\par
\textbf{Prior knowledge:} Has basic scatter-plot and line-graph reading. Lacks any data-science, modeling, or statistics background; no notion of what a ``model'' is; no functions beyond basic arithmetic.\par
\textbf{Attention span:} 20 minutes.%
}

\examplequestionbox
{typeAblue}
{typeAbackground}
{Item 2 (Type A: persona contrast)}
{%
\textbf{Topic:} Overfitting vs.\ Underfitting: Why the Best Model Is Not Always the Most Complex.\par
\textbf{Learning objective:} Deliver a clear lecture explaining the concepts of overfitting and underfitting, visually guiding students through the ``training/test error vs.\ model complexity'' curves and explicitly identifying the regions corresponding to each.\par
\textbf{Bloom-level:} \{Remember, Understand\}.\par
\textbf{Grade level:} 1st-year university student, non-CS major.\par
\textbf{Prior knowledge:} Has basic algebra and the polynomial-function concept; has heard of machine learning. Lacks hands-on ML experience; no train/test-split notion; never seen a learning curve or heard of overfitting.\par
\textbf{Attention span:} 20 minutes.%
}

\paragraph{Type B example: cognitive-level contrast.}
The two items below configure the same lesson at a lower cognitive level (Item 3) and a higher cognitive level (Item 4).

\examplequestionbox
{typeBorange}
{typeBbackground}
{Item 3 (Type B: cognitive-level contrast)}
{%
\textbf{Topic:} Why 99\% Accuracy Can Still Be a Bad Model.\par
\textbf{Learning objective:} Explain why accuracy alone can be misleading on imbalanced datasets, and describe a case where a 99\%-accurate model is still not useful.\par
\textbf{Bloom-level:} \{Remember, Understand\}.\par
\textbf{Grade level:} 10th grade.\par
\textbf{Prior knowledge:} Has percentages and ratios. Lacks confusion matrices, precision, recall, and cost-sensitive evaluation.\par
\textbf{Attention span:} 20 minutes.%
}

\examplequestionbox
{typeBorange}
{typeBbackground}
{Item 4 (Type B: cognitive-level contrast)}
{%
\textbf{Topic:} Why 99\% Accuracy Can Still Be a Bad Model.\par
\textbf{Learning objective:} Use a confusion matrix to compare accuracy, precision, and recall, and judge whether a 99\%-accurate classifier is acceptable for spam filtering, fraud detection, or disease screening.\par
\textbf{Bloom-level:} \{Remember, Understand, Apply, Analyze, Evaluate\}.\par
\textbf{Grade level:} 10th grade.\par
\textbf{Prior knowledge:} Has percentages and ratios. Lacks confusion matrices, precision, recall, and cost-sensitive evaluation.\par
\textbf{Attention span:} 20 minutes.%
}

\section{Layer 2 Crowd Pairwise Evaluation Protocol}
\label{app:layer2}

\begin{figure*}[t]
\centering
\pdfximage width \textwidth {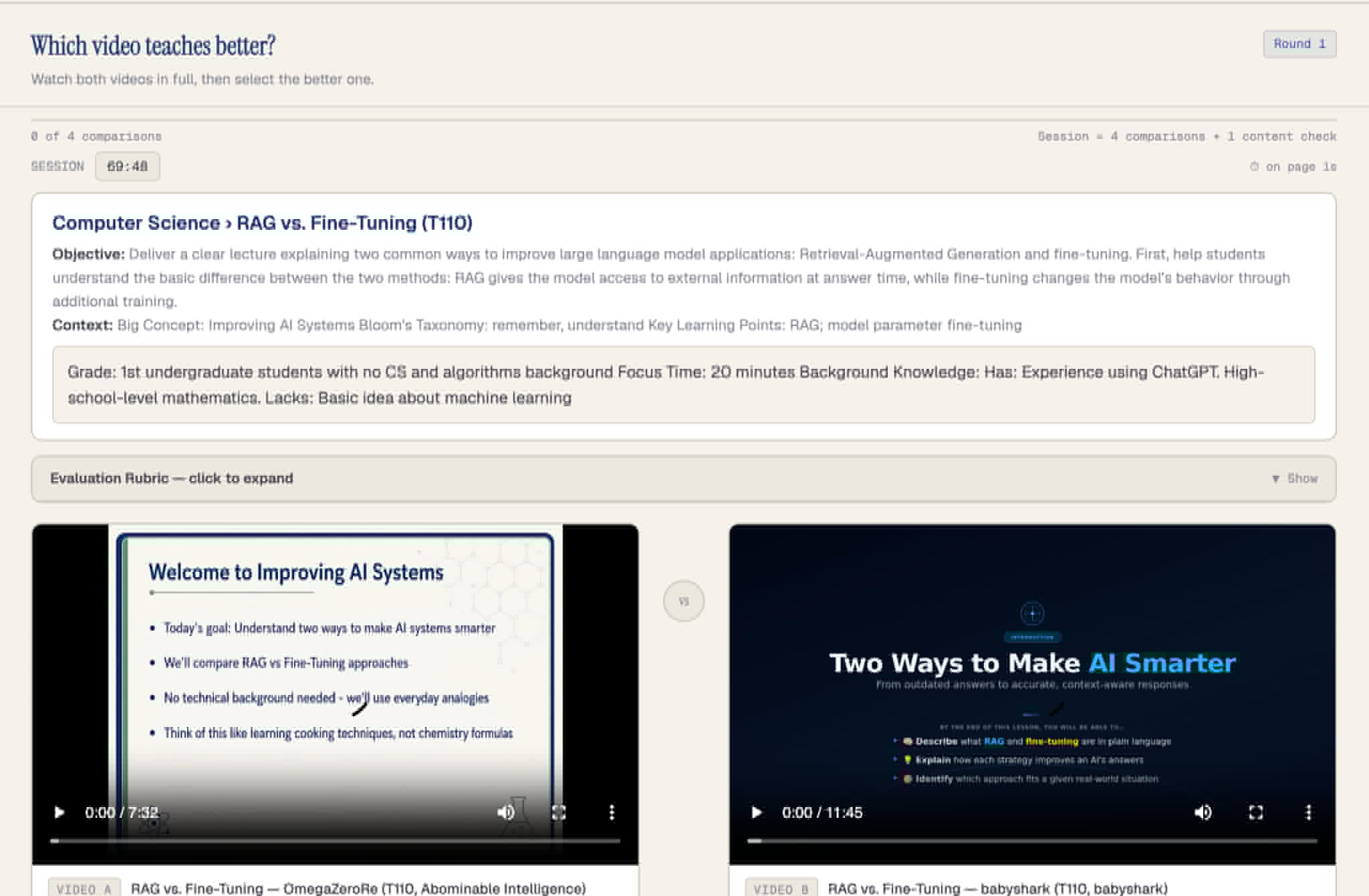}
\pdfrefximage\pdflastximage
\vspace{8pt}
\caption{Interface for the Layer 2 pairwise evaluation. Each comparison presents the learning objective, learner persona, the four dimension descriptions, and two instructional videos generated for the same item. A session contains four comparisons of the same system pair across four topics.}
\label{fig:layer2-arena}
\end{figure*}

Figure~\ref{fig:layer2-arena} illustrates the interface used for the Layer 2 pairwise evaluation. In each round, raters viewed the item information and two instructional videos generated for the same item before selecting the better teaching video.

\paragraph{Raters and Sessions.}
Raters were recruited through Prolific and screened at the platform level
for at least high-school education and English fluency. The collected corpus
comprises 246 pairwise comparisons contributed by 59 unique Prolific
participants between 25 and 31 May 2026, covering 43 of the
$\binom{10}{2}=45$ possible team-pairs. One exact duplicate
submission (identical timestamp, identifier, rationale, and vote,
from a client-side double-submit) was removed during cleaning,
leaving 246 comparisons in the analysis.
Each session locks one team-pair
for its full duration and rotates it across four distinct topics drawn
from that pair's shared topic pool, closing with a comprehension check.
Participants were compensated on Prolific at approximately \pounds 7
per hour, with a mean completion time of about 70 minutes per participant
($\approx\!\pounds 8.17$ per participant on average).

\paragraph{Attention Checks and Exclusions.}
Every session enforces four behavioural gates: (i)~a 15-second minimum on
the rubric-review page with an explicit acknowledgement checkbox; (ii)~a
per-side watch-gate that refuses submission until both videos have played
in full (2-second buffering tolerance), with forward seeks blocked on
native players and reverted on YouTube embeds; (iii)~a written
justification of at least eight characters, persisted with the vote;
(iv)~a per-session comprehension check whose failure permits joint
exclusion of the session.

\paragraph{Pair Sampling.}
Comparisons are not sampled at random. Each candidate pair $(A,B)$ is assigned a priority that combines outcome uncertainty with a rank-proximity weight, and the highest-priority pair is drawn next. Let $p = 1 / (1 + 10^{(s_B - s_A)/400})$ be the Elo-implied probability that $A$ beats $B$, so that $p(1-p)$ is largest when the two systems are closest in score, and let $m = \min(1, c/C)$ be a maturity term, where $c$ is the number of comparisons already collected in the current stage and $C$ is the total comparison budget. The evaluation runs in two stages that share $p(1-p)$ and $m$ and differ only in the rank weight:
\begin{align}
\text{Stage 1:}\quad \text{priority} &= p(1-p)\left[1 + \frac{0.35\,m}{|r_A - r_B| + 1}\right],\\
\text{Stage 2:}\quad \text{priority} &= p(1-p)\left[1 + \frac{0.8\,m}{(d_A + d_B)/2 + 0.5}\right],
\end{align}
where $r_A, r_B$ are the systems' current ranks and $d_A, d_B$ are their distances from the top-three admission boundary. To keep exposure balanced, no pair may be sampled more than two comparisons beyond the least-sampled pair. Stage~1 (190 comparisons) compares broadly to establish the field; Stage~2 (50 comparisons) concentrates on the admission boundary, for a fixed total of 246. The stage split and the weight constants ($0.35$, $0.8$) were selected by offline simulation using AI-student scores as a proxy ground truth, with the total budget fixed in advance; the stopping point is therefore independent of the observed standings.

\paragraph{Rating and Confidence Intervals.}
Each system starts at Elo 1500 and is updated after every comparison as $s \leftarrow s + 32\,(\text{outcome} - \hat{p})$, with $K=32$ and $\hat{p}$ the Elo-implied win probability. Confidence intervals come from a bootstrap that resamples the collected comparisons $B = 1{,}000$ times and recomputes Elo from scratch each time; a system's reported rating is the median over the $1{,}000$ replicates, and its 90\% interval is the 5th--95th percentile.

\paragraph{Advancement and the Third-Place Tie-Break.}
The top three systems by Elo advance. When the third slot cannot be separated within the fixed budget, it is decided by pre-specified tie-breakers rather than by extending sampling. Here \texttt{OmegaZeroRe} and \texttt{babyshark} retained overlapping 90\% intervals, so the third slot was resolved in order by (i)~head-to-head record, which was tied at 4:4; (ii)~overall Elo, $1567$ versus $1505$; and (iii)~common-opponent record against \texttt{Phd.ICU}, $60\%$ ($12$:$8$) versus $25\%$ ($2$:$6$). All three point to \texttt{OmegaZeroRe}, which therefore advanced as the third system.

\section{Layer 3 Expert Panel Evaluation Protocol}
\label{app:layer3}

The 16 final-phase items covered four AP-aligned subjects: Physics, Biology, Mathematics, and Computer Science. For each subject, the evaluation panel included two subject-matter experts and one pedagogy expert. Panel members reviewed the finalist videos using the same four evaluation dimensions described in Section~\ref{subsec:three-layer-protocol} and ranked the videos directly without computing numerical scores. The aggregated rankings determined the final challenge standing.

\begin{figure*}[t]
\centering
\pdfximage width \textwidth {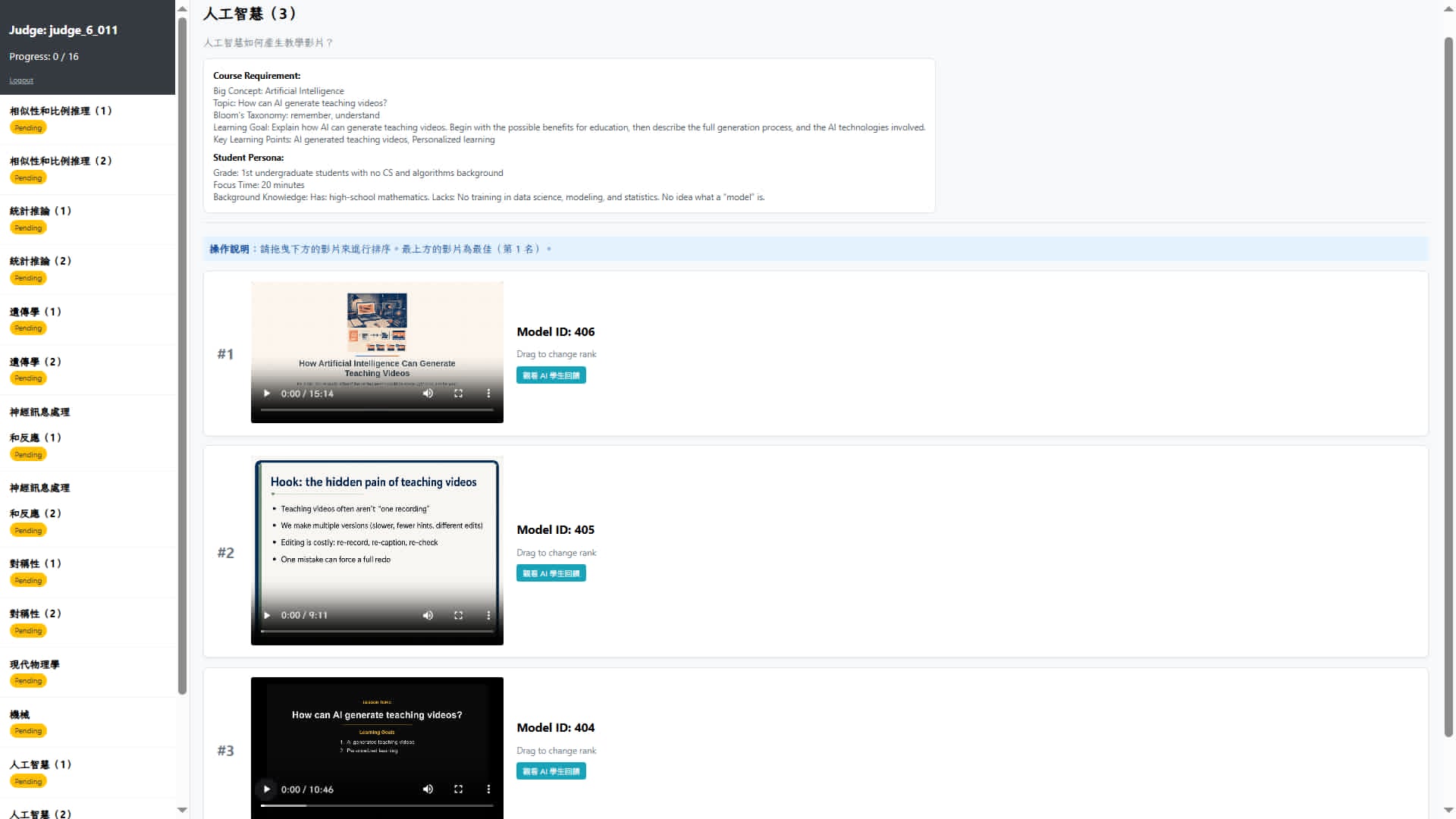}
\pdfrefximage\pdflastximage
\vspace{8pt}
\caption{Interface for the Layer 3 expert evaluation. Panel members reviewed the item information and finalist videos, then reordered the videos to record their rankings.}
\label{fig:layer3-interface}
\end{figure*}

Figure~\ref{fig:layer3-interface} shows the interface used for the Layer 3 expert evaluation. The item information was displayed above the finalist videos, which panel members reviewed and reordered to record their rankings.

\section{Participant Feedback and Survey Methodology}
\label{app:participant_feedback}
 
To contextualize the quantitative benchmark results, we analyzed post-competition surveys. The feedback survey was distributed to all submitting groups. We received 18 non-empty submissions, which were deduplicated to 15 unique team responses. Open-ended responses were coded by two annotators against a predefined codebook; Slack discussions and direct correspondence were used only as supplementary qualitative context.
 
Overall sentiment was highly positive: 13 of the 15 responding teams (approx. 87\%) valued the task as a holistic end-to-end engineering benchmark requiring tight co-optimization of planning, rendering, and narration. Four teams ($\sim 27\%$) highlighted the automated leaderboard loop, and 6 teams (40\%) praised cross-team learning. Participants treated the challenge less as a model demonstration and more as a realistic product benchmark.
 
Engineering friction, however, concentrated around the developmental iteration loop, with 11 out of 15 teams ($\sim 73\%$) reporting platform or evaluator bottlenecks. Specifically, 4 teams encountered platform UX constraints (e.g., slow score synchronization), 3 teams requested a friendlier submission API, and 5 teams asked for annotated boundary examples for rubric calibration. Crucially, 6 out of 15 teams (40\%) raised explicit concerns regarding LLM-judge noise, hallucinated feedback, and systemic mismatches with human judgments. This self-reported friction strongly corroborates the near-zero rank agreement between the automated judge and human evaluators quantified in the RQ3 analysis of the main paper.
 
The strict 30-minute latency budget effectively constrained runtime pipelines but forced explicit quality-speed trade-offs. Four teams explicitly cited latency or rate-limiting thresholds as severe bottlenecks. The primary technical barrier centered on multi-modal synchronization: 10 out of 15 teams ($\sim 67\%$) reported audio-visual misalignment, subtitle overlapping, rendering instability, or flat text-to-speech prosody. Participants' self-reports suggest that pedagogical planning is insufficient unless the final pipeline successfully synchronizes narration, visual focus, pacing, and learner needs.
 
Participant suggestions for future editions prioritize platform transparency. The highest requested feature was an inspectable evaluation trail detailing artifact hashes, fetch times, and exact video versions used for scoring. To mitigate judge variance, the pipeline should expose confidence intervals or implement score averaging. Finally, 11 of the 15 teams ($\sim 73\%$) requested higher-level capabilities (e.g., sentence-level synchronization, progressive reveals, and interactive checks), suggesting these complex dimensions be integrated as optional bonus metrics in future iterations.

\section{Persona Identification: Full Model Coefficients}
\label{app:persona-model}

The persona identification study (RQ2) asks whether raters can recover the intended learner of an instructional video.
Because each rater judges several videos and each video is judged by several raters, the 214 binary outcomes (correct / incorrect) are clustered at two levels.
Ignoring this clustering would understate the standard errors, so we fit two models that account for it in complementary ways.

\paragraph{Model 1: Variational Bayes GLMM.}
A generalized linear mixed-effects model (GLMM) with a logistic link predicts whether the rater's choice is correct from the video's group assignment (experimental, human-video, or persona-independent baseline), including crossed random intercepts for video and for participant:
\[
\mathrm{logit}\,\Pr(\text{correct}_{ij}) = \beta_0 + \beta_{\mathrm{group}} + u_{\mathrm{video}_j} + v_{\mathrm{participant}_i}.
\]
The random intercepts absorb systematic differences in item difficulty ($u$) and rater ability ($v$).
Estimation uses variational Bayes (VB), which approximates the posterior and yields posterior means with posterior standard deviations rather than frequentist $p$-values.
The estimated random-intercept standard deviations are 0.45--0.69 (video) and 0.41--0.43 (participant) on the logit scale, confirming that both clustering sources matter.

\paragraph{Model 2: GEE.}
A generalized estimating equation (GEE) with the same logistic link and the same fixed-effect predictor clusters observations by participant under an exchangeable working-correlation structure and reports robust (sandwich) standard errors.
Unlike the GLMM, GEE targets population-averaged effects and produces conventional $p$-values.

\paragraph{Reading the table.}
Table~\ref{tab:persona-coef} reports each pairwise group contrast on two scales.
The coefficient $b$ is the log-odds difference: a positive $b$ means that the first-named group has higher identification accuracy.
The odds ratio $\mathrm{OR} = e^{b}$ restates the same quantity multiplicatively: $\mathrm{OR} = 2.44$ means the odds of a correct identification are 2.44 times higher.
A 95\% confidence interval excluding 1.0 (for the OR) or 0.0 (for $b$) indicates a statistically significant difference at the 5\% level.

Both models agree on every contrast.
The experimental group's odds of correct identification are roughly 2.4 times those of the persona-independent baseline ($p = 0.007$), and the human-video group's odds are roughly 3 times the baseline's ($p = 0.002$), so both AI-adapted and human-made videos carry a learner signal that raters can perceive well above chance.
By contrast, the experimental-versus-human-video contrast is small ($\mathrm{OR} \approx 0.78$) and not significant ($p = 0.42$): raters recover the intended persona from AI-adapted videos about as often as from human-made ones, meaning the adaptation that systems produce is, in aggregate, statistically indistinguishable from that of the human topline---though this is a failure to detect a difference, not proof of equivalence.

\begin{table*}[t]
\centering
\small
\begin{tabular}{@{} l cc cc c @{}}
\toprule
 & \multicolumn{2}{c}{VB GLMM} & \multicolumn{2}{c}{GEE} & \\
\cmidrule(lr){2-3} \cmidrule(lr){4-5}
Contrast & $b$ (post.\ SD) & OR [95\% CI] & $b$ (robust SE) & OR [95\% CI] & GEE $p$ \\
\midrule
Experimental vs.\ baseline & $+0.869$ (0.189) & 2.39 [1.65, 3.46] & $+0.892$ (0.330) & 2.44 [1.28, 4.66] & 0.0069 \\
Human-video vs.\ baseline  & $+1.126$ (0.319) & 3.08 [1.65, 5.76] & $+1.092$ (0.350) & 2.98 [1.50, 5.92] & 0.0018 \\
Experimental vs.\ human-video & $-0.249$ (0.191) & 0.78 [0.54, 1.13] & $-0.244$ (0.304) & 0.78 [0.43, 1.42] & 0.4220 \\
\bottomrule
\end{tabular}
\caption{Full coefficients for the persona identification models.
Each row is one pairwise contrast between video groups.
The \emph{baseline} is the persona-independent static retrieval framework.
VB GLMM reports posterior means and posterior standard deviations; GEE reports robust standard errors and two-sided $p$-values.}
\label{tab:persona-coef}
\end{table*}

\section{Ethics Statement}
\label{app:ethics}
All human studies in this work were reviewed and approved by the behavioral and social sciences research ethics committee of the authors' institution under its minimal-risk review track.
The approval covers the Layer 2 crowd comparisons, the persona identification study, and the Layer 3 expert evaluation.

All participants were adults aged 18 or older.
No minors or members of vulnerable groups were recruited, and no participants were students or employees of the research team.
Crowd raters were recruited through Prolific.
Before each session, the platform presented the full consent document, which explains the purpose of the study, the procedure, the expected risks, and the right to withdraw at any time without penalty.
A session began only after the participant gave consent.
The expected risks were limited to mild fatigue from screen viewing.
Raters were compensated at about \pounds 7 per hour, with a mean session length of about 70 minutes.
Expert panel members likewise took part with informed consent and are reported only in aggregate.

All collected data were coded and de-identified, and identity information is stored separately from experimental data on encrypted storage that only authorized members of the research team can access.
The released human judgments contain no personal identifiers.

\end{document}